\documentclass{article} 
\usepackage{iclr2026_conference,times}

\usepackage{amsmath,amsfonts,bm}

\def\eqref#1{equation~\ref{#1}}

\def\1{\bm{1}}

\DeclareMathAlphabet{\mathsfit}{\encodingdefault}{\sfdefault}{m}{sl}
\SetMathAlphabet{\mathsfit}{bold}{\encodingdefault}{\sfdefault}{bx}{n}

\usepackage{hyperref}
\usepackage{url}
\usepackage{graphicx}
\usepackage{booktabs}
\usepackage{multirow} 
\usepackage{makecell}
\usepackage{booktabs}
\usepackage{enumitem}
\usepackage{threeparttable}
\usepackage{subcaption}
\usepackage{wrapfig}
\usepackage{tikz}

\title{MMGeoLM: Hard Negative Contrastive\\ Learning for Fine-Grained Geometric\\ Understanding in Large Multimodal Models}

\author{Kai Sun$^{*}$, Yushi Bai$^{*}$, Zhen Yang, Jiajie Zhang, Ji Qi, Lei Hou, Juanzi Li\\ 
Tsinghua University
}

\iclrfinalcopy 
\begin{document}

\maketitle

\renewcommand{\thefootnote}{\fnsymbol{footnote}}
    \footnotetext[1]{Equal Contribution.
    }
\renewcommand{\thefootnote}{\arabic{footnote}}

\begin{abstract}

Large Multimodal Models (LMMs) typically build on ViTs (e.g., CLIP), yet their training with simple random in-batch negatives limits the ability to capture fine-grained visual differences, particularly in geometric scenarios.
To address this challenge, we propose a novel \textbf{hard negative contrastive learning} framework for the vision encoder, which combines \emph{image-based contrastive learning} using generation-based hard negatives created by perturbing diagram generation code, and \emph{text-based contrastive learning} using rule-based negatives derived from modified geometric descriptions and retrieval-based negatives selected based on caption similarity.
We train a vision encoder (CLIP) using our hard negative training method, namely \texttt{MMCLIP} (Multimodal Math CLIP), and subsequently train an LMM for geometric problem-solving. Experiments show that our trained model, \texttt{MMGeoLM}, significantly outperforms other open-source models on three geometric reasoning benchmarks. Even with a size of 7B, it can rival powerful closed-source models like GPT-4o.
We further conduct ablation studies to analyze three key factors: 
hard negative types, the efficiency of image-based negatives, and training configurations.
These analyses yield important insights into optimizing the training pipeline of vision encoder for fine-grained geometric reasoning tasks.
\url{https://github.com/THU-KEG/MMGeoLM}.
\end{abstract}
\section{Introduction}

Geometric mathematical reasoning has garnered significant attention as an essential capability for large multimodal models~\citep{anthropic2024claude3,openai2023gpt4v,bai2023qwen}. It requires fine-grained identification of visual elements~\citep{lu2023mathvista} within the given images, such as geometric shapes, spatial configurations, and the relationships between them~\citep{he2024olympiadbench}. 

\begin{figure}[t!]
    \centering
    \begin{minipage}[t]{0.52\textwidth}
        \vspace{0pt}
        \centering
        \includegraphics[width=\linewidth]{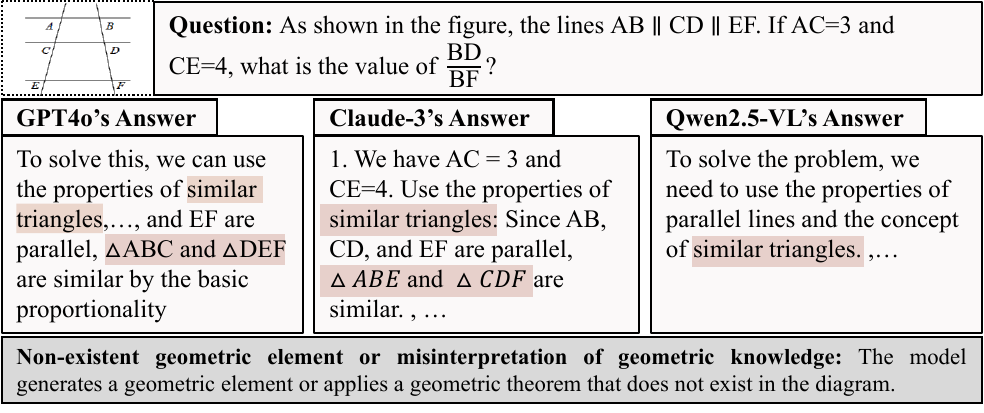}
        \caption{Examples of hallucination: non-existent elements and relation misinterpretation.}
        \label{fig:introduction}
    \end{minipage}
    \hfill
    \begin{minipage}[t]{0.46\textwidth}
        \vspace{0pt}
        \centering
        \resizebox{\linewidth}{!}{%
        \begin{tabular}{l|llc}
            \toprule
            \textbf{Model} & \textbf{Training } & \textbf{Training Data } & \textbf{MM-Math} \\
            \midrule
            AltCLIP & No    & No    & 23.8 \\
            AltCLIP & In-Batch & Randomly-sampled Neg (400K) & 24.6 \\
            \midrule
            AltCLIP & MMCLIP & \textcircled{1}: Retrieval Neg (100K) & 26.6 \\
            AltCLIP & MMCLIP & \textcircled{2}: Rule Neg (100K) & 28.1 \\
            AltCLIP & MMCLIP & \textcircled{3}: Image Neg (4K) & 28.2 \\
            \midrule
            AltCLIP & MMCLIP & \textcircled{1}+\textcircled{2} & 28.4 \\
            AltCLIP & MMCLIP & \textcircled{1}+\textcircled{2}+\textcircled{3} & \textbf{30.1} \\
            \bottomrule
        \end{tabular}
        }
        \captionof{table}{Comparison of geometric reasoning performance on MM-Math across LMMs with vision encoders trained with different hard negative data.}
        \label{tab:introduction}
    \end{minipage}
\end{figure}

However, the ``eyes'' of most existing LMMs, i.e., their pretrained vision encoders such as CLIP~\citep{patel2024tripletclip,yang2023alip,goel2022cyclip}, are primarily trained on general visual datasets that do not emphasize the intricate features necessary for specialized mathematical reasoning. Therefore, these models often fail to understand the nuanced geometric information accurately and produce incorrect reasoning and answers.
As shown in Figure~\ref{fig:introduction}, facing a simple parallel line problem, the leading LMMs such as GPT-4o~\citep{gpt4o}, Claude-3~\citep{anthropic2024claude3}, and Qwen2.5-VL~\citep{bai2025qwen2} all hallucinate non-existent elements or misinterpret spatial relationships (e.g., $\triangle ABC$, $\triangle ABE$, and the concept of similar triangles),
exhibiting notable deficiencies in capturing the intricate geometric details.

To address these shortcomings, recent research has focused on strategies such as fine-tuning on specialized mathematical datasets~\citep{gao2023g,zhang2024mavis,peng2024multimath,peng2024chimera} or utilizing massive image-caption pairs to enhance the visual perception ability of LMMs~\citep{qi2024cogcom,wang2024sam} by aligning images with corresponding captions.
However, relying solely on positive image-caption pairs may lead to spurious alignment, with models may behave like ``bag of words'' rather than achieving semantic understanding~\citep{zhang2024mathverse,doveh2023dense}.
To achieve more robust and semantically meaningful alignment, it is crucial to incorporate hard negative samples—semantically similar but mismatched pairs—which push the vision encoder to learn finer distinctions beyond shallow correlations.
Therefore, to further improve LMMs' abilities for capturing geometric information, we investigate a key question: 
\textbf{How can we systematically construct hard negatives tailored for geometric reasoning?}

In this work, we propose two types of hard negative sample construction methods, i.e., \emph{image-based} and \emph{text-based}, to enhance fine-grained geometric element recognition in vision encoder.
For text-based contrastive learning, we design two strategies to create negative captions for a geometry image: (1) a \emph{retrieval-based} method that employs dense retrieval in a geometric-domain text corpus to select captions with high lexical similarity but differing content as negative samples; and (2) a \emph{rule-based} method that modifies key geometric attributes in the captions, such as shapes, angles, and lengths, while keeping other elements unchanged, thereby producing negative samples that bear similar appearance but with key distinct information from the positives. 
For image-based contrastive learning, we introduce a novel method that leverages a large language model (LLM) to generate a detailed caption and corresponding diagram generation code for a given geometry problem, forming the positive image sample. The LLM then perturbs the code to create visually similar but geometrically incorrect diagrams, which serve as hard negative samples.
Additionally, we modify the original CLIP training framework and propose MMCLIP, a method designed to handle an arbitrary number of hard negative samples centered around a single image or caption.
Table~\ref{tab:introduction} shows the effectiveness of MMCLIP training on different hard negative sets.

We evaluate our trained MMGeoLM  on four geometric benchmarks spanning two categories: choice-based questions (GeoQA, MathVista, and We-Math) and open-ended questions (MM-Math).
Results show that the proposed model outperforms all existing open-source models on GeoQA and MathVista, and achieves state-of-the-art performance on MM-Math, surpassing GPT-4o by 7.4\%.
Ablation studies confirm both text-based and image-based hard negatives benefit model performance. Notably, using only 4K image-based negatives yields better results than 100K retrieval-based ones. These findings demonstrate that our hard negative construction and training strategy significantly enhances reasoning accuracy in geometric reasoning.
\begin{figure*}[!ht]
    \centering
    \includegraphics[width=\linewidth]{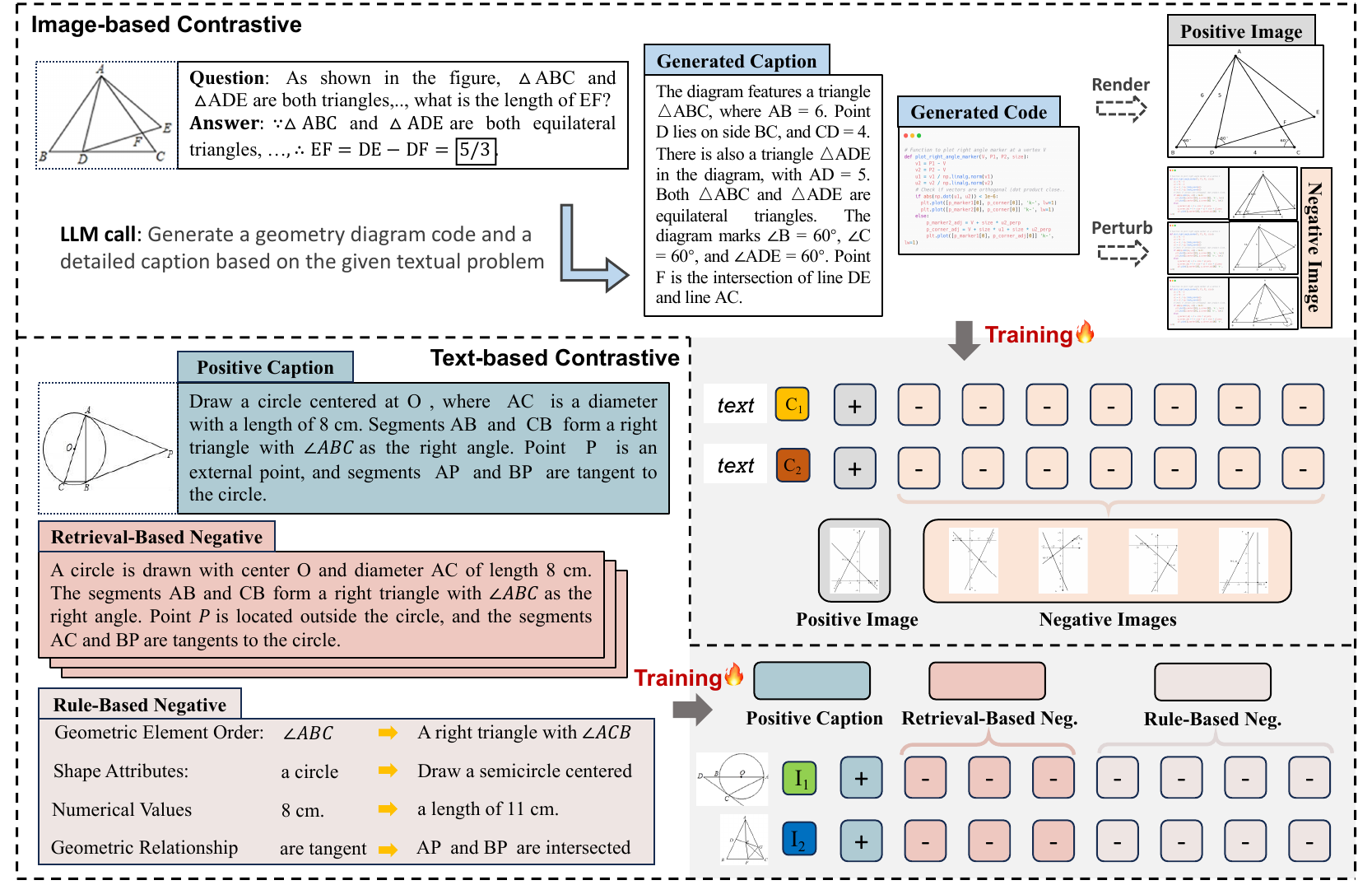}
    \caption{Image-based and text-based hard negative construction and the corresponding MMCLIP training method.}
    \label{fig:negative}
\end{figure*}

\section{Hard Negative Data Construction}

\subsection{Negative Images Construction}
\label{section:negative_image_section}

We construct  negative geometric examples by generating geometric diagram code for positive images and perturbing it to create negatives. 
Existing generation methods~\citep{zhang2024mavis,zou2024dynamath,wei2024slow} rely on handcrafted rules that cover limited elements and overlook inter-element relationships, resulting in a gap from real-world problems. 
In contrast, authentic geometry problems~\citep{sun2024mm} are more diverse and accurate, with geometric elements described either explicitly (e.g., “AB is 8 cm”) or implicitly through reasoning (e.g., using the Pythagorean theorem), and reflected in the corresponding geometric diagrams~\citep{mcclintock2002students}.
Benefiting from advancements in mathematical reasoning and code-generation capabilities of LLMs~\citep{openai2024o1}, which can effectively parse geometric conditions, perform necessary reasoning, and generate executable code to produce geometric diagrams closely matching the original images. As illustrated in Figure~\ref{fig:negative}, providing the LLM with both the problem and the answer yields code and a caption. Executing the code produces diagrams that not only replicate the original figures but may also include additional \textit{numeric markings}.

To handle occasional errors in model-generated code, we adopt a lightweight model~\citep{glm2024chatglm} to correct syntax issues when the code fails.
Subsequently, we generate detailed captions directly from code using Gemini 2.5 Pro~\citep{google2025gemini}.
Upon evaluating 123 generated captions, three independent annotators confirmed that the accuracy of captions in representing geometric elements reaches 100\%. Further details are provided in Appendix~\ref{sec:hunman verification}.

To create the negative images, we design prompts for LLMs to generate 10 negative captions for each positive caption, differing subtly but critically. We then employ Gemini 2.5 Pro to modify the Python scripts based on these perturbed captions. These modified scripts produce diagrams that closely resemble the originals while aligning with the intended constraints. Representative examples of the constructed image-based negatives are provided in Appendix~\ref{sec:case of image_based negatives}.

\subsection{Negative Captions Construction}
\label{section:negative_caption}
We propose two methods for constructing text-based hard negative samples. The first method retrieves captions similar to each positive example from a large image-caption dataset. The second method generates targeted hard negative samples based on reasoning errors observed during LMMs' inference. 

\paragraph{Retrieval-based Hard Negative.}
Many previous works in open-domain QA take the top-ranked instances recalled by the retriever as negative examples to further improve retriever performance~\citep{karpukhin2020dense,xiong2020approximate,huang2020embedding}. However, they suffer from the risks of introducing false negatives, which means some related instances are incorrectly treated as negatives~\citep{xiong2020approximate,zhou2022simans,yang2024trisampler}.
To address this issue, we build upon the recently proposed Mavis dataset~\citep{zhang2024mavis}, which mitigates false negatives by ensuring a one-to-one correspondence between each image and its caption.
We use SimANS~\citep{zhou2022simans} to encode all captions, compute pairwise similarities, and retrieve the top 100 most similar captions per image as hard negatives.

\paragraph{Rule-based Hard Negative.}
Existing studies modify positive captions by randomly altering nouns, attributes, and relationships~\citep{yuksekgonul2022and,zhang2024contrasting}, while others leverage in-context learning to generate meaningful modifications~\citep{patel2025tripletclip}. However, these methods either introduce semantic ambiguity or focus solely on the vision encoder level, without considering downstream reasoning errors, as exemplified by the misunderstanding in Figure~\ref{fig:introduction}, where parallel lines are mistaken as forming a triangle.

To address this issue, we analyze the evaluation results of LMMs on MM-Math~\citep{sun2024mm} and identify four major types of image element recognition errors: 

\begin{enumerate}[itemsep=0pt, leftmargin=*, itemsep=0pt,partopsep=0pt,parsep=0pt]
    \item \emph{Geometric element ordering}: Modifying the sequence of alphabetical order in geometric diagrams while ensuring the new order does not match the original cyclically (e.g., $ABCD$ changing to $CDAB$ is invalid).  
    \item \emph{Shape attributes}: Altering properties such as changing squares to rectangles or right triangles to isosceles triangles.  
    \item \emph{Geometric relationships}: Modifying relationships such as parallelism between two lines or similarity between two triangles.  
    \item \emph{Numerical values}: Adjusting numerical values in captions, such as modifying angles or segment lengths.  
\end{enumerate}

To address these error types, we design rule-based strategies that use GLM-4~\citep{glm2024chatglm} to modify each positive caption and generate 10 hard negative examples.
These text-based negatives are used in model training, and two representative types are shown in Figure~\ref{fig:negative}.

\section{MMGeoLM: Architecture and Training}
\begin{figure*}[!t]
    \centering
    \small
    \includegraphics[width=\linewidth]{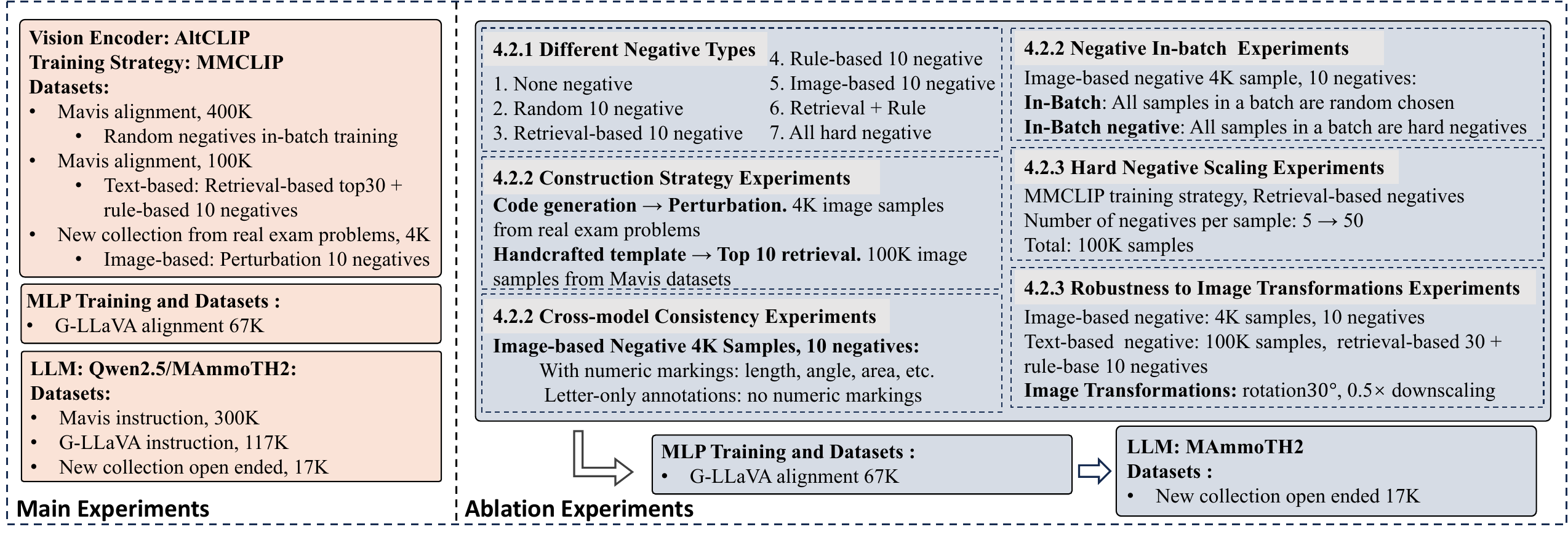}
    \caption{Overview of the MMGeoLM training pipeline, including main and each ablation experiment configurations, training strategies, and datasets.}
    \label{fig:training_pipeline}
\end{figure*}

\subsection{Architecture}
We adopt the LLaVA architecture~\citep{liu2024improved}, which comprises three components: (1) a vision encoder, (2) a 2-layer MLP adapter, and (3) an LLM backbone. For the LLM backbone, we use MAmmoTH2-7B~\citep{yue2024mammoth2} and Qwen2.5-7B-Instruct~\citep{qwen2025qwen25technicalreport}. The vision encoder is based on AltCLIP~\citep{chen2022altclip}, configured with a maximum length of 512 tokens and a model size of 0.5B parameters.

\subsection{Training Pipeline}
Our training pipeline, illustrated in Figure~\ref{fig:training_pipeline}, consists of three stages.
\textbf{(1)} We first train CLIP with our MMCLIP training strategy detailed in Section~\ref{section:trainging strategy}. For text-based hard negatives, we adopt the Mavis image-caption alignment dataset~\citep{zhang2024mavis}, while for image-based negatives, we collect 4K geometry questions from real middle-school exams. Additional details, including the negative types and negative-to-positive ratio, are provided in Figure~\ref{fig:training_pipeline}, and further statistics across categories of hard negatives are reported in Appendix~\ref{sec:hard_negative_statistics}.
\textbf{(2)} In the second stage, we fine-tune the MLP adapter using 67K image-text pairs from the G-LLaVA dataset~\citep{gao2023g}.
\textbf{(3)} In the third stage, we conduct supervised fine-tuning (SFT). We construct a high-quality training dataset of 17K geometry problems from the \texttt{21st Century Education}\footnote{\url{https://www.21cnjy.com/}}. This dataset aligns with middle school curricula and reflects real-world student assessment scenarios.
Each problem includes a detailed analysis, from which we extract the core solution to create a structured step-by-step reasoning process. The final answers are enclosed in \texttt{\textbackslash boxed\{\}}, enabling targeted learning. 
For our main experiments, we use a combined SFT dataset comprising 300K Mavis instruction data~\citep{zhang2024mavis}, 117K G-LLaVA instruction data~\citep{gao2023g}, and the 17K collected open-ended geometry questions. For ablation studies, we use only the 17K geometry problems for efficient validation.
We refer to the final trained model as \texttt{MMGeoLM}.

\subsection{Hard Negative CLIP Training}

\label{section:trainging strategy}

\textbf{In-batch Training}.
Conventional CLIP-trained vision encoders~\citep{radford2021learning,doveh2023teaching} typically combine random examples for a batch and adopt \emph{in-batch training loss over all the samples} in each batch. For each batch containing samples $\{(I_i, T_i)_{i=1}^N\}$, where each \(I_i, T_i\) represents a pair of image and text caption, the loss is calculated as

\begin{equation}
    \mathcal{L} = \frac{1}{N}\sum_{i=1}^{N}\left[-\ln \frac{\exp(s(I_i, T_i))}{\sum_j \exp(s(I_i, T_j))}-\ln \frac{\exp(s(I_i, T_i))}{\sum_j \exp(s(I_j, T_i))}\right],
\end{equation}
where \( \text{s}(\cdot, \cdot) \) denotes the similarity function over the feature encodings. 

\textbf{MMCLIP Training}.
In our data construction process, hard negative captions and images are generated for each image-caption pair without providing corresponding positive samples for these hard negatives. As a result, our MMCLIP training derives the \emph{loss solely from one positive sample and its associated hard negatives} within each batch.
This guides the model to discern subtle differences, enhancing its fine-grained geometric understanding.
Formally, given a text-based negative training batch $\{I, T^+,T_i^-|_{i=1}^{N}\}$, where \( I \) represents the image, \( T^+ \) denotes the positive caption and \( T_i^- \) is the negative caption conducted around the image. The loss function is formulated as follows:

\begin{equation}
\mathcal{L} =  -\ln \frac{\exp(s(I,T^+))}{\exp(s(I,T^+))+\sum_{i=1}^{N} \exp(s(I,T_i^-))}.
\label{eq:clip_loss}
\end{equation}
Since negative samples are constructed around elements within the image, their score \( s(I,T_i^-)\) is expected to be higher than a randomly chosen one. From the above equation, we calculate the gradient for positive and negative samples:

\begin{figure}[!h]
\centering
\begin{minipage}{0.44\linewidth}
\begin{align*}
\frac{\partial \mathcal{L}}{\partial s^+} 
&= - \frac{\sum_{i=1}^{N} \exp(s^-_i)}
{\exp(s^+) + \sum_{i=1}^{N} \exp(s^-_i)}
\end{align*}
\end{minipage}%
\begin{minipage}{0.44\linewidth}
\begin{align*}
\frac{\partial \mathcal{L}}{\partial s^-_i} 
&= \frac{\exp(s^-_i)}
{\exp(s^+) + \sum_{i=1}^{N} \exp(s^-_i)}
\end{align*}
\end{minipage}
\end{figure}

where $s^+=s(I,T^+)$ and $s_i^-=s(I,T_i^-)$. Since the number of hard negatives is unrestricted, the gradient functions can be effectively optimized, leading to improved image-caption alignment. 
The training strategy is illustrated in Figure~\ref{fig:negative}, where we also provide image-based negatives training batches $\{T, I^+,I_i^-|_{i=1}^{M}\}$, the loss can be derived similarly to Equation~\ref{eq:clip_loss}.
Unlike in-batch training, which relies on random batch-internal negatives, MMCLIP allows training with independently constructed negatives, making it suitable for  unimodal negatives, such as rule-based or retrieval-based. We further validate  different training strategies only for hard negatives in the ablation experiments.

\section{Experiments}

To quantitatively evaluate the effectiveness of our MMGeoLM in geometric problem-solving, we conduct comparative experiments on four geometric benchmarks using various LMMs. Additionally, we analyze the impact of different vision encoder types, image-based negatives and training strategies in the subsequent ablation studies.

\subsection{Main Experiment: MMGeoLM Evaluation}
\subsubsection{Datasets and Metrics}
We assess the geometric reasoning capabilities of various LMMs across two types of benchmarks: multiple-choice and open-ended. The multiple-choice benchmark includes GeoQA~\citep{chen2021geoqa}, a geometric question answering task based on plane geometry; We-Math~\citep{qiao2024we}, a visual mathematical reasoning task with questions of varying difficulty, solvable in two or three steps; and MathVista~\citep{lu2023mathvista}, widely used for evaluating LMMs' performance. The open-ended benchmark is MM-Math~\citep{sun2024mm}, which features high discriminative difficulty sourced from secondary school-level problems. We use accuracy (ACC) as the metric. For We-Math, we assess multi-step problems (S2 and S3), while for MM-Math, we evaluate models across easy, medium, and hard categories using a test set of 700 problems. For MathVista, we evaluate geometry problem-solving (GEO) and algebraic (ALG) categories.
\subsubsection{Baselines}
The evaluated LMMs are categorized into two groups: Closed-source APIs: Claude-3-Opus~\citep{anthropic2024claude3}, Claude-3.5-Sonnet, GPT-4o-20240513~\citep{gpt4o}, and GPT-4V. Open-source LMMs: MAVIS-7B~\citep{zhang2024mavis}, G-LLaVA-7B~\citep{gao2023g}, InternVL2-8B~\citep{chen2024expanding}, LLaVA-OneVision-7B~\citep{li2024llava}, 
Qwen2.5-VL-7B-Instruct~\citep{bai2025qwen2}, Chimera-Reasoner-8B~\citep{peng2024chimera}, InternLM-XComposer2-7B~\citep{dong2024internlm}, Phi-3-Vision-128K-Instruct~\citep{abdin2024phi}, and Math-LLaVA-13B~\citep{shi2024math}. For the open-source category, we choose LMMs that have achieved strong results on geometric benchmarks~\citep{chen2021geoqa,lu2023mathvista}. 
Additionally, we include human evaluation baselines, using scores from prior studies~\citep{zhang2024mavis,sun2024mm}.  

\subsubsection{Overall Results}

\begin{table*}[!t]
\centering
\resizebox{\linewidth}{!}{%
\begin{tabular}{l|c|cc|cc|cccc}
\toprule
\multirow{2}{*}{\textbf{Model}} & \multirow{2}{*}{GeoQA} &
\multicolumn{2}{c|}{MathVista} & \multicolumn{2}{c|}{We-Math} & 
\multicolumn{4}{c}{MM-Math} \\
\cmidrule(lr){3-4}\cmidrule(lr){5-6}\cmidrule(lr){7-10}
& & GEO & ALG & S2 & S3 & Easy & Med & Hard & Avg \\
\midrule
Human* & 92.3 & 48.4 & 50.9 & - & - & 90.7 & 81.9 & 47.6 & 80.4 \\
\midrule
Claude-3-Opus*      & 44.5 & 46.3 & 46.6 & 32.9 & 23.0 & 29.5 & 19.3 & 3.6 & 20.3 \\
Claude-3.5-Sonnet   & 65.1 & 66.3 & 68.4 & \textbf{64.7} & \textbf{62.1} & 34.4 & 31.9 & 13.6 & 31.7 \\
GPT-4V*             & -    & 50.5 & 53.0 & 49.2 & 38.2 & 37.8 & 21.2 & 1.8 & 23.1 \\
GPT-4o  & 58.9 & 62.7 & 65.4 & 58.0 & 43.6 & 45.8 & 30.0 & 10.9 & 31.8 \\
\midrule
MAVIS-7B*                 & 68.3 & 64.1 & 59.2 & 37.9 & 34.6 & - & - & - & - \\
G-LLaVA-7B*               & 67.0 & 56.7 & -    & 30.1 & 32.7 & - & - & - & - \\
Math-LLaVA-13B            & 48.4 & 55.6 & 55.0 & 31.7 & 23.0 & - & - & - & - \\
InternVL2-8B              & 56.4 & 60.5 & 60.8 & 41.1 & 37.1 & 33.6 & 21.7 & 9.0 & 23.3 \\
LLaVA-OneVision-7B        & 64.6 & 54.6 & 53.1 & 28.7 & 22.3 & 40.5 & 24.8 & 10.5 & 27.0 \\
Chimera-Reasoner-8B       & \textbf{69.6} & 48.5 & 42.6 & 29.9 & 31.6 & 36.2 & 22.2 & 9.0 & 24.1 \\
Phi-3-Vision-128K-Instruct  & 29.5 & 38.8 & 40.0 & 33.3 & 30.1 & 12.9 & 7.1 & 0.0 & 7.8 \\
Qwen2.5-VL-7B-Instruct      & 59.0 & 67.7 & 66.9 & 58.1 & 50.6 & 50.7 & 32.2 & \textbf{14.2} & 34.8 \\
InternLM-XComposer2-7B    & 38.8 & 63.0 & 56.6 & 33.1 & 33.0 & 18.9 & 12.2 & 4.5 & 13.1 \\
\midrule
MMGeoLM-MAmmoTH2-7B        & 68.5 & 68.7 & \textbf{69.5} & 41.5 & 37.5 & 52.5 & 32.4 & 4.5 & 36.9 \\
\quad \emph{w/o MMCLIP}           & 55.4 & 52.2 & 51.3 & 34.1 & 35.3 & 42.5 & 21.8 & 4.5 & 26.8 \\
\quad \emph{Original AltCLIP}     & 46.7 & 45.8 & 46.8 & 30.1 & 29.2 & 38.2 & 20.4 & 1.6 & 25.9 \\
\addlinespace[0.2em]      
MMGeoLM-Qwen2.5-7B         & 69.2 & \textbf{69.8} & 68.1 & 39.8 & 36.4 & \textbf{55.3} & \textbf{36.9} & 9.0 & \textbf{39.2} \\
\quad \emph{w/o MMCLIP}           & 56.3 & 54.1 & 50.3 & 35.2 & 35.1 & 50.5 & 28.8 & 4.5 & 34.3 \\
\quad \emph{Original AltCLIP}     & 50.7 & 47.6 & 48.9 & 33.1 & 32.7 & 52.5 & 30.8 & 4.5 & 33.7 \\
\bottomrule
\end{tabular}}
\caption{Accuracy (\%) of various LMMs on GeoQA, MathVista, We-Math, and MM-Math. `S2/S3' denotes the two-step settings in We-Math. Results marked * are taken from their original papers.}
\label{tab:main}
\end{table*}

As shown in Table~\ref{tab:main}, our proposed MMGeoLM achieves state-of-the-art performance on the MathVista and MM-Math benchmarks. On the GeoQA benchmark, MMGeoLM-Qwen2.5-7B lags 0.4\% behind Chimera-Reasoner-8B. As Chimera-Reasoner-8B was trained on GeoQA~\citep{peng2024chimera}, MMGeoLM-Qwen2.5-7B achieves the best performance among other models that were not trained on this dataset. The improved performance in geometric problem-solving demonstrates the effectiveness of our training approach.
For the We-Math benchmark, MMGeoLM-MAmmoTH2-7B underperforms GPT-4o, Claude-3.5-Sonnet, and the open-source Qwen2.5-VL-7B-Instruct model. We attribute this to We-Math's emphasis on recognition from visual elements rather than geometric mathematical reasoning, which limits the effectiveness of MMGeoLM-MAmmoTH2-7B's geometric reasoning enhancement strategies.
On the open-ended MM-Math benchmark, MMGeoLM-Qwen2.5-7B performs poorly on hard problems, with less than 10\% accuracy, but achieves over 55\% accuracy on easy problems. Easy problems require fewer reasoning steps, allowing geometric element recognition to significantly enhance accuracy. In contrast, hard problems involve multi-step reasoning, where its impact is more limited. 
Compared to using the \emph{w/o MMCLIP} (In-batch trained AltCLIP with random negatives), the vision encoder trained with our constructed hard negative samples significantly enhances MMGeoLM’s geometric understanding capabilities. To rule out the possibility that performance gains stem from training–test leakage, we perform a contamination analysis (Appendix~\ref{sec:training-test overlap}) and we confirm that the evaluation sets contain no near-duplicate items.

\subsection{Ablations}

\subsubsection{Ablation Study I: Different Negative Types}

\begin{table}[!ht]
\centering
\resizebox{0.7\linewidth}{!}{%
\begin{tabular}{llccccc}
\toprule
\multirow{2}{*}{$\text{Vision Encoder}$} & \multirow{2}{*}{\#Num.} &\multirow{2}{*}{GeoQA} &
\multicolumn{4}{c}{MM-MATH }\\
\cmidrule(l){4-7}
 & & & Easy & Med & Hard & Avg \\
\midrule
{\color[HTML]{7F7F7F} Original}                            & {\color[HTML]{7F7F7F} -}    & {\color[HTML]{7F7F7F} 45.4} & {\color[HTML]{7F7F7F} 40.5} & {\color[HTML]{7F7F7F} 20.8} & {\color[HTML]{7F7F7F} 4.5} & {\color[HTML]{7F7F7F} 23.8} \\
{\color[HTML]{7F7F7F} $\text{Random}_{\text{10}}$}       & {\color[HTML]{7F7F7F} 400K} & {\color[HTML]{7F7F7F} 45.0} & {\color[HTML]{7F7F7F} 39.8} & {\color[HTML]{7F7F7F} 22.7} & {\color[HTML]{7F7F7F} 4.5} & {\color[HTML]{7F7F7F} 24.9} \\
$\text{Retrieval}_{\text{10}}$         & 100K           & 53.0 & 43.6 & 24.8 & 4.5 & 26.6 \\
$\text{Rule}_{\text{10}}$        & 100K           & 48.8 & 45.4 & 27.6 & 4.5 & 28.1 \\
$\text{Image-based}_{\text{10}}$         & 4K             & 54.9 & 45.6 & 26.5 & 4.5 & 29.0 \\
\midrule
$\text{Retrieval}_{\text{10}}$+$\text{Rule}_{\text{10}}$   & 200K           & 56.5 & 47.5 & 28.3 & 4.5 & 29.4 \\
\textbf{All Negatives} & 204K           & \textbf{58.2} & \textbf{49.8} & \textbf{29.1} & 4.5 & \textbf{30.4} \\
\bottomrule
\end{tabular}
}
\caption{Ablation study on vision encoder training with different types and quantities of hard negatives. The bottom panel reports performance when combining multiple types.}
\label{tab:ablation_combined}
\end{table}

To further evaluate the performance of vision encoders trained with different types of hard negatives, we compare seven negatives summarized in Table~\ref{tab:ablation_combined}. All models are based on AltCLIP. 
The \textcolor{gray}{Original} vision encoder is directly initialized without any additional training, while \textcolor{gray}{$\text{Random}_{\text{10}}$} follows the conventional in-batch random negatives sampling strategy in Section~\ref{section:trainging strategy}. 
The remaining five vision encoders are trained with our proposed MMCLIP method in Section~\ref{section:trainging strategy}. The subscript `10' indicates a 1:10 positive-to-negative ratio. We further complement this analysis with retrieval evaluations reported in Appendix~\ref{sec:retieval}, using the Hit@1 metric on four constructed evaluation sets.

\textbf{Results}.
As shown in Table~\ref{tab:ablation_combined}, vision encoders trained with hard negatives consistently outperform random in-batch negatives. 
even with only 4K image-based negatives , the $\text{Image-based}_{\text{10}}$ encoder achieves the highest average accuracy (29.0\%) among all single hard negative methods on MM-Math. 
Furthermore, while $\text{Retrieval}_{10}$ and $\text{Rule}_{10}$  negatives individually provide moderate gains, their combination further improves performance (29.4\%), indicating their complementary nature.
The best  results of  All Negatives are achieved when all three types of negatives are combined, demonstrating that diverse hard negatives collectively enhance the model's geometric understanding. These findings underscore the importance of both the \textit{type} and \textit{diversity} of hard negatives in contrastive training for vision encoders.

\subsubsection{Ablation Study II: Image-based Negatives}

\begin{table}[t!]
    \centering
    \begin{minipage}[t]{0.48\textwidth}
        \centering

        \vspace{-0.2cm}
        \footnotesize
        \setlength{\tabcolsep}{3.2pt}
        \begin{tabular}{@{}lcccccc@{}}
        \toprule
        \multirow{2}{*}{Method} & \multirow{2}{*}{\#Num} & \multirow{2}{*}{GeoQA} & \multicolumn{4}{c}{MM-Math} \\
        \cmidrule(l){4-7}
         & &  & Easy & Med & Hard &Avg \\
        \midrule
        \textsc{$\text{Rule}_{img}$}      & 100K & 52.9 & 44.0 & 25.4 & 1.0 & 28.3\\
        \textsc{$\text{Perturb}$}  & 4K   & 54.9 & 45.6 & 26.5 & 4.5 &29.0\\
        \textbf{$\Delta$} & \textbf{-96K} & \textbf{+2.0} & \textbf{+1.6} & \textbf{+1.1} & \textbf{+3.5} &\textbf{+0.7}\\
        \bottomrule
        \end{tabular}
                \caption{Effect of different image negative construction strategies. $\Delta$ rows give absolute gains of \textsc{Perturb} over $\text{Rule}_{img}$.}
        \label{tab:img_source_ablation}
    \end{minipage}
    \hfill
    \begin{minipage}[t]{0.48\textwidth}
        \centering

        \vspace{-0.2cm}
        \footnotesize
        \setlength{\tabcolsep}{2.5pt}
        \begin{tabular}{@{}lccccc@{}}
        \toprule
        \multirow{2}{*}{Visual Marking} & \multirow{2}{*}{GeoQA} & \multicolumn{4}{c}{MM-MATH} \\
        \cmidrule(l){3-6}
        & & Easy & Med & Hard & Avg \\
        \midrule
        \textbf{w/} length/angle/etc. & 54.9 & 45.6 & 26.5 & 4.5 & 29.0 \\
        \textbf{w/o} length/angle/etc. & 51.7 & 50.0 & 25.1 & 4.5 & 28.6 \\
        \bottomrule
        \end{tabular}
                \caption{Impact of numeric markings on vision encoder training. \textbf{w/} and \textbf{w/o} denote settings with and without such numeric annotations, respectively.}
        \label{tab:annotation_mm}
    \end{minipage}
\end{table}

\textbf{Code Perturbation vs. Rule-design: Assessing Image Negatives Construction Methods}.
We contrast two image–based negative construction methods. 
\textsc{Perturb} applies our code perturbation method to 4K images, generating negative images that preserve similar structure while introducing subtle differences. In contrast, $\text{Rule}_{img}$ is built from manually designed templates in the Mavis corpus. We construct 100K samples and retrieve the most similar images as hard negatives. To ensure a fair comparison, we fix the positive-to-negative ratio at 1:10 in both settings.
Results in Table~\ref{tab:img_source_ablation} show that the 4K \textsc{Perturb} setting exceeds the 100K $\text{Rule}_{img}$ on both GeoQA (54.9 vs.\ 52.9) and MM‑Math (29.0 vs.\ 28.3). 
These results highlight that code-perturbed negatives, despite being smaller in scale, outperform synthetic data due to their semantic diversity and closer resemblance to real-world tasks. This advantage is corroborated by additional analyses in Appendix~\ref{sec:4kimage_based}, where \textsc{Perturb}-based negatives are shown to have a more distinct distribution than $\text{Rule}_{img}$ counterparts and higher semantic similarity to real-world supervised data.

\begin{table}[t!]
    \centering
    \begin{minipage}[t]{0.48\textwidth}
        \centering

        \vspace{-0.2cm}
        \includegraphics[width=\linewidth]{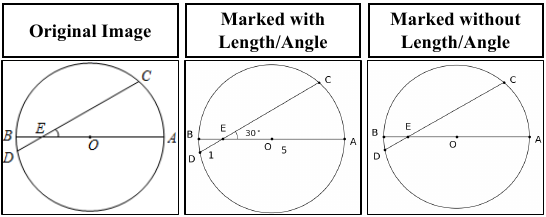}
                \captionof{figure}{An example of vision encoder training data with or without numeric markings.}
        \label{fig:marked_angle}
        \vspace{-3mm}
    \end{minipage}
    \hfill
    \begin{minipage}[t]{0.48\textwidth}
        \centering

        \vspace{-0.2cm}
        \footnotesize
        \setlength{\tabcolsep}{2.8pt}
        \begin{tabular}{@{}lccccc@{}}
        \toprule
        \multirow{2}{*}{In-Batch Negative} & \multirow{2}{*}{GeoQA} & \multicolumn{4}{c}{MM-MATH} \\
        \cmidrule(l){3-6}
         & & Easy & Med & Hard & Avg \\
        \midrule
        Random & 46.2 & 41.8 & 23.7 & 4.5 & 25.5 \\
        Image-based NS & \textbf{57.3} & \textbf{51.7}  & \textbf{29.0}  & 4.5 & \textbf{32.1} \\
        \bottomrule
        \end{tabular}
        \caption{Comparison of in-batch training with random negatives vs. in-batch training with image-based negatives. Mean score over 5 random seeds on 4K image-based negatives.}
        \label{tab:inbatch_vs_mmclip}
    \end{minipage}
\end{table}

\textbf{Length/Angle Marked Matters? Cross‑Modal Consistency Analysis}.
For a large amount of geometric problems, the conditions (lengths, angles, etc.) are provided only in textual format without corresponding image markings. 
This inherent asymmetry between image and text may lead to cross-modal hallucinations (vision encoder training vs. LMMs training).
To assess these inconsistencies, we remove all numeric markings (lengths, angles,  etc.) from the image-based negatives, retaining only letter labels (see Figure~\ref{fig:marked_angle}) for the vision encoder.
Results in Table~\ref{tab:annotation_mm} show that without numeric markings leads to a slight performance drop (e.g., 29.0\% $\rightarrow$ 28.6\% on Avg, 54.9\%$\rightarrow$ 51.7\% on GeoQA), suggesting that numeric markings have a moderate but non-negligible impact on alignment learning. 
This shows that the model primarily learns to align geometric features with text through structural cues, while numeric markings serve as auxiliary signals rather than essential supervision.

\textbf{Hard Negative Performance Gains or Trade-off for In-Batch Training}.
We leverage the proposed image-based negative strategy for the in-batch training to verify the effectiveness. 
In the in-batch negatives setting, each sample in the batch serves as a hard negative for others, with traditional in-batch typically selected at random.
For a fair comparison, both methods use 10 negative examples per positive sample.
As shown in Table~\ref{tab:inbatch_vs_mmclip}, in-batch training with our proposed image-based negative strategy outperforms In-Batch on the \textit{easy}, \textit{medium}, and \textit{average} subsets under the same amount of training data. This highlights the advantage of explicitly constructed image-based negatives in guiding the model to better distinguish visually similar examples, thus achieving higher overall accuracy. Furthermore, these results indicate that carefully selected mutual hard negatives within batches lead to more effective contrastive learning.

\subsubsection{Ablation Study III: Training Strategy and Robustness}

\begin{table*}[t!]
    \centering
    \begin{minipage}[t]{0.48\textwidth}
        \centering

        \vspace{-0.2cm}
        \includegraphics[width=\linewidth]{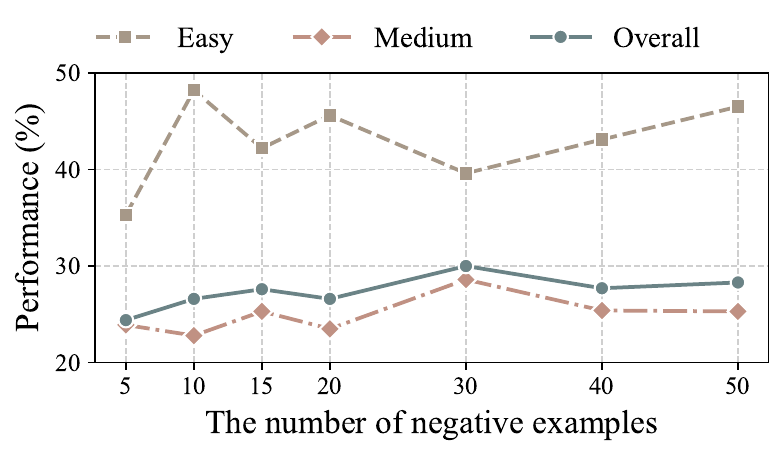}
                \captionof{figure}{MMGeoLM performance with varying numbers of hard negative ratio.}
        \label{fig:num strong negative}

        \vspace{-3mm}
    \end{minipage}
    \hfill
    \begin{minipage}[t]{0.48\textwidth}
        \centering

        \vspace{-0.2cm}
        \small
        \setlength{\tabcolsep}{2.8pt}
        \begin{tabular}{@{}lccccc@{}}
        \toprule
        \multirow{2}{*}{Transformers} & \multirow{2}{*}{Negatives}  & \multicolumn{4}{c}{MM-MATH} \\ 
        \cmidrule(l){3-6}
        & & Easy & Med & Hard & Avg \\
        \midrule
        No-Changing    & Text-based   & 48.6 & 29.1 & 9.0 & 31.9 \\
        Rotate$30^\circ$    & Text-based    & 47.4 & 28.5 & 9.0 & 31.0 \\
        Downscaling      & Text-based     & 45.6 & 26.8 & 0.0 & 28.4 \\
        \midrule
        No-Changing    & Image-based & 45.6 & 26.5 & 4.5 & 29.0 \\
        Rotate$30^\circ$    & Image-based & 50.0 & 25.6 & 4.5 & 29.0 \\
        Downscaling      & Image-based & 48.2 & 25.6 & 4.5 & 28.7 \\
        \bottomrule
        \end{tabular}
        \caption{Robustness evaluation of MMGeoLM trained on text/image negatives under image transformers (rotation and downscaling).}
        \label{tab:performance}
    \end{minipage}
\end{table*}

\textbf{Data Scaling: More Hard Negatives Are Not Always Better}.
Our proposed MMGeoLM method allows flexible scaling of negatives. However, whether performance improves proportionally with an increasing number of hard negatives remains unclear. To address this, we investigate the impact of varying the number of retrieval-based negatives per sample from 5 to 50. 
 As shown in Figure~\ref{fig:num strong negative}, MMGeoLM's performance on MM-Math improves with increasing hard negative samples from 5 to 30, but slightly declines beyond this point. These results indicate that  hard negative samples have diminishing returns beyond a certain threshold, with excessive examples reducing model performance.  

\textbf{Robustness to Geometric Image Transformations}.
Diagrams in real exams often undergo transformations such as rotation or scaling. To evaluate MMGeoLM's robustness, we apply two image perturbations: (1) a $30^\circ$ clockwise rotation and (2) downscaling by 0.5. Table~\ref{tab:performance} reports the results under both \textit{text-based} and \textit{image-based} negative conditions. 
Model performance remains largely stable under a $30^\circ$ rotation, showing only a minor accuracy drop of 0.9\% in the text-based setting. This suggests the model is robust against moderate image rotations. In contrast, downscaling significantly impacts accuracy, dropping text-based performance from 31.9\% to 28.4\%. This decline indicates that textual captions are particularly sensitive to reduced image resolution whereas the image-based negative demonstrates stronger robustness, benefiting from diagram scale variations.

\section{Related Work}
Mathematics-related research has recently received significant attention in large models. For text-only mathematical reasoning, several works employ external tools like Tora~\citep{gou2023tora} or intermediate step decomposition methods such as MAmmoTH~\citep{yue2023mammoth}, Metamath~\citep{yu2023metamath} and Math-Shepherd~\citep{yue2023mammoth}. Many multimodal mathematics benchmarks have been proposed, including mathematical competition-oriented OlympiadBench~\citep{he2024olympiadbench}, geometry-focused datasets like Geometry3K~\citep{lu2021inter}, VisScience~\citep{jiang2024visscience}, UniGeo~\citep{chen2022unigeo} and GeoQA~\citep{chen2022unigeo}. Many works take effort on training multimodal math, including  G-llava~\citep{gao2023g}, Meta-LLaVA~\citep{shi2024math} and MAVIS~\citep{zhang2024mavis}. These models often struggle with fine-grained recognition of geometric elements. 
Vision encoders are critical for capturing spatial and structural information in multimodal tasks. Prior works~\citep{zhang2024contrasting,doveh2023dense,doveh2023teaching,singh2023coarse} enhance image understanding using negative captions, but rarely target fine-grained geometric recognition. Methods such as NegCLIP~\citep{yuksekgonul2022and} and TriCLIP~\citep{yang2024trisampler} focus on learning from negative samples, yet their effectiveness is limited by small or randomly selected negatives. In this work, we propose a scalable and geometry-aware negative learning strategy that explicitly constructs task-relevant hard negatives to improve geometric understanding.
\section{Conclusion}
This paper introduces two types of hard negatives, image-based and text-based negatives, targeted at geometric element understanding to enhance LMMs' geometric reasoning. Our resulting model, MMGeoLM, significantly outperformed existing models, even surpassing GPT-4o on key benchmarks.
Our ablation analyses corroborate three key findings. 
First, both image-based and text-based negatives enhance geometric problem-solving, and their effects are complementary when used jointly. 
Second, the proposed image-based negatives, constructed via code perturbation, exhibit advantages in terms of data efficiency, annotation robustness, and  enhanced performance combined with traditional in-batch training methods. 
Third, increasing the number of negatives does not monotonically improve performance; instead, our MMCLIP training strategy demonstrates robustness by balancing the quality and diversity of negatives, yielding consistent gains across benchmarks.


\section*{Ethics Statement}
Our 4K image-based negatives and 17K supervised fine-tuning training data were collected from the \texttt{21st Century Education}. We have obtained an official authorization agreement from them for research.

\section*{Reproducibility Statement}

The training dataset, 4K image-based negatives and 17K supervised fine-tuning training data, will be publicly released after the paper publication, with a download link to be provided in the camera-ready version.
For the reproducibility of our training strategy, we provide three components. 
First, the implementation of MMCLIP training with both image-based and text-based negatives, along with the corresponding hyperparameters, is included in the supplementary material. 
Second, the MLP training setup follows the official implementation released in the LLaVA GitHub repository. 
Third, for incorporating LLMs, we adopt the training code and parameters from the official LLaVA GitHub repository for MAmmoTH2-7B, and we use the official code and configurations provided in the LLaVA-Next GitHub repository for Qwen2.5-7B.

\section*{The Use of Large Language Models}

In this work, large language models were employed in three ways. 
First, they were used to generate code-based geometric diagrams and the corresponding captions for constructing image-based negatives. 
Second, they were applied to modify positive captions in order to derive negative captions for text-based negatives. 
Third, they were utilized to assist in polishing the writing of the manuscript.

\bibliography{iclr2026_conference,custom}

\begin{thebibliography}{53}
\providecommand{\natexlab}[1]{#1}
\providecommand{\url}[1]{\texttt{#1}}
\expandafter\ifx\csname urlstyle\endcsname\relax
  \providecommand{\doi}[1]{doi: #1}\else
  \providecommand{\doi}{doi: \begingroup \urlstyle{rm}\Url}\fi

\bibitem[Abdin et~al.(2024)Abdin, Aneja, Awadalla, Awadallah, Awan, Bach, Bahree, Bakhtiari, Bao, Behl, et~al.]{abdin2024phi}
Marah Abdin, Jyoti Aneja, Hany Awadalla, Ahmed Awadallah, Ammar~Ahmad Awan, Nguyen Bach, Amit Bahree, Arash Bakhtiari, Jianmin Bao, Harkirat Behl, et~al.
\newblock Phi-3 technical report: A highly capable language model locally on your phone.
\newblock \emph{arXiv preprint arXiv:2404.14219}, 2024.

\bibitem[Anthropic(2024)]{anthropic2024claude3}
Anthropic.
\newblock {Claude3} system card, 2024.
\newblock URL \url{https://www.anthropic.com/news/claude-3-family}.

\bibitem[Bai et~al.(2023)Bai, Bai, Yang, Wang, Tan, Wang, Lin, Zhou, and Zhou]{bai2023qwen}
Jinze Bai, Shuai Bai, Shusheng Yang, Shijie Wang, Sinan Tan, Peng Wang, Junyang Lin, Chang Zhou, and Jingren Zhou.
\newblock Qwen-vl: A frontier large vision-language model with versatile abilities.
\newblock \emph{arXiv preprint arXiv:2308.12966}, 2023.

\bibitem[Bai et~al.(2025)Bai, Chen, Liu, Wang, Ge, Song, Dang, Wang, Wang, Tang, et~al.]{bai2025qwen2}
Shuai Bai, Keqin Chen, Xuejing Liu, Jialin Wang, Wenbin Ge, Sibo Song, Kai Dang, Peng Wang, Shijie Wang, Jun Tang, et~al.
\newblock Qwen2. 5-vl technical report.
\newblock \emph{arXiv preprint arXiv:2502.13923}, 2025.

\bibitem[Chen et~al.(2021)Chen, Tang, Qin, Liang, Liu, Xing, and Lin]{chen2021geoqa}
Jiaqi Chen, Jianheng Tang, Jinghui Qin, Xiaodan Liang, Lingbo Liu, Eric~P Xing, and Liang Lin.
\newblock Geoqa: A geometric question answering benchmark towards multimodal numerical reasoning.
\newblock \emph{arXiv preprint arXiv:2105.14517}, 2021.

\bibitem[Chen et~al.(2022{\natexlab{a}})Chen, Li, Qin, Lu, Lin, Chen, and Liang]{chen2022unigeo}
Jiaqi Chen, Tong Li, Jinghui Qin, Pan Lu, Liang Lin, Chongyu Chen, and Xiaodan Liang.
\newblock Unigeo: Unifying geometry logical reasoning via reformulating mathematical expression.
\newblock \emph{arXiv preprint arXiv:2212.02746}, 2022{\natexlab{a}}.

\bibitem[Chen et~al.(2024)Chen, Wang, Cao, Liu, Gao, Cui, Zhu, Ye, Tian, Liu, et~al.]{chen2024expanding}
Zhe Chen, Weiyun Wang, Yue Cao, Yangzhou Liu, Zhangwei Gao, Erfei Cui, Jinguo Zhu, Shenglong Ye, Hao Tian, Zhaoyang Liu, et~al.
\newblock Expanding performance boundaries of open-source multimodal models with model, data, and test-time scaling.
\newblock \emph{arXiv preprint arXiv:2412.05271}, 2024.

\bibitem[Chen et~al.(2022{\natexlab{b}})Chen, Liu, Zhang, Ye, Yang, and Wu]{chen2022altclip}
Zhongzhi Chen, Guang Liu, Bo-Wen Zhang, Fulong Ye, Qinghong Yang, and Ledell Wu.
\newblock Altclip: Altering the language encoder in clip for extended language capabilities.
\newblock \emph{arXiv preprint arXiv:2211.06679}, 2022{\natexlab{b}}.

\bibitem[DeepMind(2025)]{google2025gemini}
Google DeepMind.
\newblock Gemini 2.5: Our most intelligent ai model.
\newblock \url{https://blog.google/technology/google-deepmind/gemini-model-thinking-updates-march-2025/}, 2025.
\newblock Accessed: 2025-05-19.

\bibitem[Dong et~al.(2024)Dong, Zhang, Zang, Cao, Wang, Ouyang, Wei, Zhang, Duan, Cao, et~al.]{dong2024internlm}
Xiaoyi Dong, Pan Zhang, Yuhang Zang, Yuhang Cao, Bin Wang, Linke Ouyang, Xilin Wei, Songyang Zhang, Haodong Duan, Maosong Cao, et~al.
\newblock Internlm-xcomposer2: Mastering free-form text-image composition and comprehension in vision-language large model.
\newblock \emph{arXiv preprint arXiv:2401.16420}, 2024.

\bibitem[Doveh et~al.(2023{\natexlab{a}})Doveh, Arbelle, Harary, Herzig, Kim, Cascante-Bonilla, Alfassy, Panda, Giryes, Feris, et~al.]{doveh2023dense}
Sivan Doveh, Assaf Arbelle, Sivan Harary, Roei Herzig, Donghyun Kim, Paola Cascante-Bonilla, Amit Alfassy, Rameswar Panda, Raja Giryes, Rogerio Feris, et~al.
\newblock Dense and aligned captions (dac) promote compositional reasoning in vl models.
\newblock \emph{Advances in Neural Information Processing Systems}, 36:\penalty0 76137--76150, 2023{\natexlab{a}}.

\bibitem[Doveh et~al.(2023{\natexlab{b}})Doveh, Arbelle, Harary, Schwartz, Herzig, Giryes, Feris, Panda, Ullman, and Karlinsky]{doveh2023teaching}
Sivan Doveh, Assaf Arbelle, Sivan Harary, Eli Schwartz, Roei Herzig, Raja Giryes, Rogerio Feris, Rameswar Panda, Shimon Ullman, and Leonid Karlinsky.
\newblock Teaching structured vision \& language concepts to vision \& language models.
\newblock In \emph{Proceedings of the IEEE/CVF Conference on Computer Vision and Pattern Recognition}, pp.\  2657--2668, 2023{\natexlab{b}}.

\bibitem[Gao et~al.(2023)Gao, Pi, Zhang, Ye, Zhong, Wang, Hong, Han, Xu, Li, et~al.]{gao2023g}
Jiahui Gao, Renjie Pi, Jipeng Zhang, Jiacheng Ye, Wanjun Zhong, Yufei Wang, Lanqing Hong, Jianhua Han, Hang Xu, Zhenguo Li, et~al.
\newblock G-llava: Solving geometric problem with multi-modal large language model.
\newblock \emph{arXiv preprint arXiv:2312.11370}, 2023.

\bibitem[GLM et~al.(2024)GLM, Zeng, Xu, Wang, Zhang, Yin, Zhang, Rojas, Feng, Zhao, et~al.]{glm2024chatglm}
Team GLM, Aohan Zeng, Bin Xu, Bowen Wang, Chenhui Zhang, Da~Yin, Dan Zhang, Diego Rojas, Guanyu Feng, Hanlin Zhao, et~al.
\newblock Chatglm: A family of large language models from glm-130b to glm-4 all tools.
\newblock \emph{arXiv preprint arXiv:2406.12793}, 2024.

\bibitem[Goel et~al.(2022)Goel, Bansal, Bhatia, Rossi, Vinay, and Grover]{goel2022cyclip}
Shashank Goel, Hritik Bansal, Sumit Bhatia, Ryan Rossi, Vishwa Vinay, and Aditya Grover.
\newblock Cyclip: Cyclic contrastive language-image pretraining.
\newblock \emph{Advances in Neural Information Processing Systems}, 35:\penalty0 6704--6719, 2022.

\bibitem[Gou et~al.(2023)Gou, Shao, Gong, Shen, Yang, Huang, Duan, and Chen]{gou2023tora}
Zhibin Gou, Zhihong Shao, Yeyun Gong, Yelong Shen, Yujiu Yang, Minlie Huang, Nan Duan, and Weizhu Chen.
\newblock Tora: A tool-integrated reasoning agent for mathematical problem solving.
\newblock \emph{arXiv preprint arXiv:2309.17452}, 2023.

\bibitem[He et~al.(2024)He, Luo, Bai, Hu, Thai, Shen, Hu, Han, Huang, Zhang, et~al.]{he2024olympiadbench}
Chaoqun He, Renjie Luo, Yuzhuo Bai, Shengding Hu, Zhen~Leng Thai, Junhao Shen, Jinyi Hu, Xu~Han, Yujie Huang, Yuxiang Zhang, et~al.
\newblock Olympiadbench: A challenging benchmark for promoting agi with olympiad-level bilingual multimodal scientific problems.
\newblock \emph{arXiv preprint arXiv:2402.14008}, 2024.

\bibitem[Huang et~al.(2020)Huang, Sharma, Sun, Xia, Zhang, Pronin, Padmanabhan, Ottaviano, and Yang]{huang2020embedding}
Jui-Ting Huang, Ashish Sharma, Shuying Sun, Li~Xia, David Zhang, Philip Pronin, Janani Padmanabhan, Giuseppe Ottaviano, and Linjun Yang.
\newblock Embedding-based retrieval in facebook search.
\newblock In \emph{Proceedings of the 26th ACM SIGKDD International Conference on Knowledge Discovery \& Data Mining}, pp.\  2553--2561, 2020.

\bibitem[Jiang et~al.(2024)Jiang, Yang, Chen, Du, Wang, Xu, and Tang]{jiang2024visscience}
Zhihuan Jiang, Zhen Yang, Jinhao Chen, Zhengxiao Du, Weihan Wang, Bin Xu, and Jie Tang.
\newblock Visscience: An extensive benchmark for evaluating k12 educational multi-modal scientific reasoning.
\newblock \emph{arXiv preprint arXiv:2409.13730}, 2024.

\bibitem[Karpukhin et~al.(2020)Karpukhin, O{\u{g}}uz, Min, Lewis, Wu, Edunov, Chen, and Yih]{karpukhin2020dense}
Vladimir Karpukhin, Barlas O{\u{g}}uz, Sewon Min, Patrick Lewis, Ledell Wu, Sergey Edunov, Danqi Chen, and Wen-tau Yih.
\newblock Dense passage retrieval for open-domain question answering.
\newblock \emph{arXiv preprint arXiv:2004.04906}, 2020.

\bibitem[Li et~al.(2024)Li, Zhang, Guo, Zhang, Li, Zhang, Zhang, Zhang, Li, Liu, et~al.]{li2024llava}
Bo~Li, Yuanhan Zhang, Dong Guo, Renrui Zhang, Feng Li, Hao Zhang, Kaichen Zhang, Peiyuan Zhang, Yanwei Li, Ziwei Liu, et~al.
\newblock Llava-onevision: Easy visual task transfer.
\newblock \emph{arXiv preprint arXiv:2408.03326}, 2024.

\bibitem[Liu et~al.(2024)Liu, Li, Li, and Lee]{liu2024improved}
Haotian Liu, Chunyuan Li, Yuheng Li, and Yong~Jae Lee.
\newblock Improved baselines with visual instruction tuning.
\newblock In \emph{Proceedings of the IEEE/CVF Conference on Computer Vision and Pattern Recognition}, pp.\  26296--26306, 2024.

\bibitem[Lu et~al.(2021)Lu, Gong, Jiang, Qiu, Huang, Liang, and Zhu]{lu2021inter}
Pan Lu, Ran Gong, Shibiao Jiang, Liang Qiu, Siyuan Huang, Xiaodan Liang, and Song-Chun Zhu.
\newblock Inter-gps: Interpretable geometry problem solving with formal language and symbolic reasoning.
\newblock \emph{arXiv preprint arXiv:2105.04165}, 2021.

\bibitem[Lu et~al.(2023)Lu, Bansal, Xia, Liu, Li, Hajishirzi, Cheng, Chang, Galley, and Gao]{lu2023mathvista}
Pan Lu, Hritik Bansal, Tony Xia, Jiacheng Liu, Chunyuan Li, Hannaneh Hajishirzi, Hao Cheng, Kai-Wei Chang, Michel Galley, and Jianfeng Gao.
\newblock Mathvista: Evaluating mathematical reasoning of foundation models in visual contexts.
\newblock \emph{arXiv preprint arXiv:2310.02255}, 2023.

\bibitem[McClintock et~al.(2002)McClintock, Jiang, and July]{mcclintock2002students}
Edwin McClintock, Zhonghong Jiang, and Raquel July.
\newblock Students' development of three-dimensional visualization in the geometer's sketchpad environment.
\newblock 2002.

\bibitem[OpenAI(2023)]{openai2023gpt4v}
OpenAI.
\newblock {GPT-4V(ision)} system card, 2023.
\newblock URL \url{https://openai.com/research/gpt-4v-system-card}.

\bibitem[OpenAI(2024{\natexlab{a}})]{gpt4o}
OpenAI.
\newblock Hello gpt-4o.
\newblock \url{https://openai.com/index/hello-gpt-4o/}, 2024{\natexlab{a}}.

\bibitem[OpenAI(2024{\natexlab{b}})]{openai2024o1}
OpenAI.
\newblock Openai o1 system card.
\newblock \url{https://openai.com/index/openai-o1-system-card/}, 2024{\natexlab{b}}.
\newblock Accessed: 2025-05-19.

\bibitem[Patel et~al.(2024)Patel, Kusumba, Cheng, Kim, Gokhale, Baral, and Yang]{patel2024tripletclip}
Maitreya Patel, Abhiram Kusumba, Sheng Cheng, Changhoon Kim, Tejas Gokhale, Chitta Baral, and Yezhou Yang.
\newblock Tripletclip: Improving compositional reasoning of clip via synthetic vision-language negatives.
\newblock \emph{arXiv preprint arXiv:2411.02545}, 2024.

\bibitem[Patel et~al.(2025)Patel, Kusumba, Cheng, Kim, Gokhale, Baral, et~al.]{patel2025tripletclip}
Maitreya Patel, Naga Sai~Abhiram Kusumba, Sheng Cheng, Changhoon Kim, Tejas Gokhale, Chitta Baral, et~al.
\newblock Tripletclip: Improving compositional reasoning of clip via synthetic vision-language negatives.
\newblock \emph{Advances in Neural Information Processing Systems}, 37:\penalty0 32731--32760, 2025.

\bibitem[Peng et~al.(2024{\natexlab{a}})Peng, Fu, Gao, Zhong, Fu, and Tang]{peng2024multimath}
Shuai Peng, Di~Fu, Liangcai Gao, Xiuqin Zhong, Hongguang Fu, and Zhi Tang.
\newblock Multimath: Bridging visual and mathematical reasoning for large language models.
\newblock \emph{arXiv preprint arXiv:2409.00147}, 2024{\natexlab{a}}.

\bibitem[Peng et~al.(2024{\natexlab{b}})Peng, Li, Zhou, Xia, Zhang, Bai, Mao, Wang, He, Zhou, et~al.]{peng2024chimera}
Tianshuo Peng, Mingsheng Li, Hongbin Zhou, Renqiu Xia, Renrui Zhang, Lei Bai, Song Mao, Bin Wang, Conghui He, Aojun Zhou, et~al.
\newblock Chimera: Improving generalist model with domain-specific experts.
\newblock \emph{arXiv preprint arXiv:2412.05983}, 2024{\natexlab{b}}.

\bibitem[Qi et~al.(2024)Qi, Ding, Wang, Bai, Lv, Hong, Xu, Hou, Li, Dong, et~al.]{qi2024cogcom}
Ji~Qi, Ming Ding, Weihan Wang, Yushi Bai, Qingsong Lv, Wenyi Hong, Bin Xu, Lei Hou, Juanzi Li, Yuxiao Dong, et~al.
\newblock Cogcom: Train large vision-language models diving into details through chain of manipulations.
\newblock \emph{arXiv preprint arXiv:2402.04236}, 2024.

\bibitem[Qiao et~al.(2024)Qiao, Tan, Dong, Wu, Sun, Song, GongQue, Lei, Wei, Zhang, et~al.]{qiao2024we}
Runqi Qiao, Qiuna Tan, Guanting Dong, Minhui Wu, Chong Sun, Xiaoshuai Song, Zhuoma GongQue, Shanglin Lei, Zhe Wei, Miaoxuan Zhang, et~al.
\newblock We-math: Does your large multimodal model achieve human-like mathematical reasoning?
\newblock \emph{arXiv preprint arXiv:2407.01284}, 2024.

\bibitem[Qwen et~al.(2025)Qwen, :, Yang, Yang, Zhang, Hui, Zheng, Yu, Li, Liu, Huang, Wei, Lin, Yang, Tu, Zhang, Yang, Yang, Zhou, Lin, Dang, Lu, Bao, Yang, Yu, Li, Xue, Zhang, Zhu, Men, Lin, Li, Tang, Xia, Ren, Ren, Fan, Su, Zhang, Wan, Liu, Cui, Zhang, and Qiu]{qwen2025qwen25technicalreport}
Qwen, :, An~Yang, Baosong Yang, Beichen Zhang, Binyuan Hui, Bo~Zheng, Bowen Yu, Chengyuan Li, Dayiheng Liu, Fei Huang, Haoran Wei, Huan Lin, Jian Yang, Jianhong Tu, Jianwei Zhang, Jianxin Yang, Jiaxi Yang, Jingren Zhou, Junyang Lin, Kai Dang, Keming Lu, Keqin Bao, Kexin Yang, Le~Yu, Mei Li, Mingfeng Xue, Pei Zhang, Qin Zhu, Rui Men, Runji Lin, Tianhao Li, Tianyi Tang, Tingyu Xia, Xingzhang Ren, Xuancheng Ren, Yang Fan, Yang Su, Yichang Zhang, Yu~Wan, Yuqiong Liu, Zeyu Cui, Zhenru Zhang, and Zihan Qiu.
\newblock Qwen2.5 technical report, 2025.
\newblock URL \url{https://arxiv.org/abs/2412.15115}.

\bibitem[Radford et~al.(2021)Radford, Kim, Hallacy, Ramesh, Goh, Agarwal, Sastry, Askell, Mishkin, Clark, et~al.]{radford2021learning}
Alec Radford, Jong~Wook Kim, Chris Hallacy, Aditya Ramesh, Gabriel Goh, Sandhini Agarwal, Girish Sastry, Amanda Askell, Pamela Mishkin, Jack Clark, et~al.
\newblock Learning transferable visual models from natural language supervision.
\newblock In \emph{International conference on machine learning}, pp.\  8748--8763. PMLR, 2021.

\bibitem[Shi et~al.(2024)Shi, Hu, Bin, Liu, Yang, Ng, Bing, and Lee]{shi2024math}
Wenhao Shi, Zhiqiang Hu, Yi~Bin, Junhua Liu, Yang Yang, See-Kiong Ng, Lidong Bing, and Roy Ka-Wei Lee.
\newblock Math-llava: Bootstrapping mathematical reasoning for multimodal large language models.
\newblock \emph{arXiv preprint arXiv:2406.17294}, 2024.

\bibitem[Singh et~al.(2023)Singh, Zhang, Wang, Wang, Xiong, Du, and Chen]{singh2023coarse}
Harman Singh, Pengchuan Zhang, Qifan Wang, Mengjiao Wang, Wenhan Xiong, Jingfei Du, and Yu~Chen.
\newblock Coarse-to-fine contrastive learning in image-text-graph space for improved vision-language compositionality.
\newblock \emph{arXiv preprint arXiv:2305.13812}, 2023.

\bibitem[Sun et~al.(2024)Sun, Bai, Qi, Hou, and Li]{sun2024mm}
Kai Sun, Yushi Bai, Ji~Qi, Lei Hou, and Juanzi Li.
\newblock Mm-math: Advancing multimodal math evaluation with process evaluation and fine-grained classification.
\newblock In \emph{Findings of the Association for Computational Linguistics: EMNLP 2024}, pp.\  1358--1375, 2024.

\bibitem[Wang et~al.(2024)Wang, Vasu, Faghri, Vemulapalli, Farajtabar, Mehta, Rastegari, Tuzel, and Pouransari]{wang2024sam}
Haoxiang Wang, Pavan Kumar~Anasosalu Vasu, Fartash Faghri, Raviteja Vemulapalli, Mehrdad Farajtabar, Sachin Mehta, Mohammad Rastegari, Oncel Tuzel, and Hadi Pouransari.
\newblock Sam-clip: Merging vision foundation models towards semantic and spatial understanding.
\newblock In \emph{Proceedings of the IEEE/CVF Conference on Computer Vision and Pattern Recognition}, pp.\  3635--3647, 2024.

\bibitem[Wei et~al.(2024)Wei, Yin, Li, Wang, Zhao, Sun, Ge, Zhang, and Jiang]{wei2024slow}
Haoran Wei, Youyang Yin, Yumeng Li, Jia Wang, Liang Zhao, Jianjian Sun, Zheng Ge, Xiangyu Zhang, and Daxin Jiang.
\newblock Slow perception: Let's perceive geometric figures step-by-step.
\newblock \emph{arXiv preprint arXiv:2412.20631}, 2024.

\bibitem[Xiong et~al.(2020)Xiong, Xiong, Li, Tang, Liu, Bennett, Ahmed, and Overwijk]{xiong2020approximate}
Lee Xiong, Chenyan Xiong, Ye~Li, Kwok-Fung Tang, Jialin Liu, Paul Bennett, Junaid Ahmed, and Arnold Overwijk.
\newblock Approximate nearest neighbor negative contrastive learning for dense text retrieval.
\newblock \emph{arXiv preprint arXiv:2007.00808}, 2020.

\bibitem[Yang et~al.(2023)Yang, Deng, An, Li, Feng, Guo, Yang, and Liu]{yang2023alip}
Kaicheng Yang, Jiankang Deng, Xiang An, Jiawei Li, Ziyong Feng, Jia Guo, Jing Yang, and Tongliang Liu.
\newblock Alip: Adaptive language-image pre-training with synthetic caption.
\newblock In \emph{Proceedings of the IEEE/CVF International Conference on Computer Vision}, pp.\  2922--2931, 2023.

\bibitem[Yang et~al.(2024)Yang, Shao, Dong, and Tang]{yang2024trisampler}
Zhen Yang, Zhou Shao, Yuxiao Dong, and Jie Tang.
\newblock Trisampler: A better negative sampling principle for dense retrieval.
\newblock In \emph{Proceedings of the AAAI Conference on Artificial Intelligence}, volume~38, pp.\  9269--9277, 2024.

\bibitem[Yu et~al.(2023)Yu, Jiang, Shi, Yu, Liu, Zhang, Kwok, Li, Weller, and Liu]{yu2023metamath}
Longhui Yu, Weisen Jiang, Han Shi, Jincheng Yu, Zhengying Liu, Yu~Zhang, James~T Kwok, Zhenguo Li, Adrian Weller, and Weiyang Liu.
\newblock Metamath: Bootstrap your own mathematical questions for large language models.
\newblock \emph{arXiv preprint arXiv:2309.12284}, 2023.

\bibitem[Yue et~al.(2023)Yue, Qu, Zhang, Fu, Huang, Sun, Su, and Chen]{yue2023mammoth}
Xiang Yue, Xingwei Qu, Ge~Zhang, Yao Fu, Wenhao Huang, Huan Sun, Yu~Su, and Wenhu Chen.
\newblock Mammoth: Building math generalist models through hybrid instruction tuning.
\newblock \emph{arXiv preprint arXiv:2309.05653}, 2023.

\bibitem[Yue et~al.(2024)Yue, Zheng, Zhang, and Chen]{yue2024mammoth2}
Xiang Yue, Tuney Zheng, Ge~Zhang, and Wenhu Chen.
\newblock Mammoth2: Scaling instructions from the web.
\newblock \emph{arXiv preprint arXiv:2405.03548}, 2024.

\bibitem[Yuksekgonul et~al.(2022)Yuksekgonul, Bianchi, Kalluri, Jurafsky, and Zou]{yuksekgonul2022and}
Mert Yuksekgonul, Federico Bianchi, Pratyusha Kalluri, Dan Jurafsky, and James Zou.
\newblock When and why vision-language models behave like bags-of-words, and what to do about it?
\newblock \emph{arXiv preprint arXiv:2210.01936}, 2022.

\bibitem[Zhang et~al.(2024{\natexlab{a}})Zhang, Awal, and Agrawal]{zhang2024contrasting}
Le~Zhang, Rabiul Awal, and Aishwarya Agrawal.
\newblock Contrasting intra-modal and ranking cross-modal hard negatives to enhance visio-linguistic compositional understanding.
\newblock In \emph{Proceedings of the IEEE/CVF Conference on Computer Vision and Pattern Recognition}, pp.\  13774--13784, 2024{\natexlab{a}}.

\bibitem[Zhang et~al.(2024{\natexlab{b}})Zhang, Jiang, Zhang, Lin, Guo, Qiu, Zhou, Lu, Chang, Qiao, et~al.]{zhang2024mathverse}
Renrui Zhang, Dongzhi Jiang, Yichi Zhang, Haokun Lin, Ziyu Guo, Pengshuo Qiu, Aojun Zhou, Pan Lu, Kai-Wei Chang, Yu~Qiao, et~al.
\newblock Mathverse: Does your multi-modal llm truly see the diagrams in visual math problems?
\newblock In \emph{European Conference on Computer Vision}, pp.\  169--186. Springer, 2024{\natexlab{b}}.

\bibitem[Zhang et~al.(2024{\natexlab{c}})Zhang, Wei, Jiang, Guo, Li, Zhang, Tong, Liu, Zhou, Wei, et~al.]{zhang2024mavis}
Renrui Zhang, Xinyu Wei, Dongzhi Jiang, Ziyu Guo, Shicheng Li, Yichi Zhang, Chengzhuo Tong, Jiaming Liu, Aojun Zhou, Bin Wei, et~al.
\newblock Mavis: Mathematical visual instruction tuning with an automatic data engine.
\newblock \emph{arXiv preprint arXiv:2407.08739}, 2024{\natexlab{c}}.

\bibitem[Zhou et~al.(2022)Zhou, Gong, Liu, Zhao, Shen, Dong, Lu, Majumder, Wen, Duan, et~al.]{zhou2022simans}
Kun Zhou, Yeyun Gong, Xiao Liu, Wayne~Xin Zhao, Yelong Shen, Anlei Dong, Jingwen Lu, Rangan Majumder, Ji-Rong Wen, Nan Duan, et~al.
\newblock Simans: Simple ambiguous negatives sampling for dense text retrieval.
\newblock \emph{arXiv preprint arXiv:2210.11773}, 2022.

\bibitem[Zou et~al.(2024)Zou, Guo, Yang, Zhang, Hu, and Zhang]{zou2024dynamath}
Chengke Zou, Xingang Guo, Rui Yang, Junyu Zhang, Bin Hu, and Huan Zhang.
\newblock Dynamath: A dynamic visual benchmark for evaluating mathematical reasoning robustness of vision language models.
\newblock \emph{arXiv preprint arXiv:2411.00836}, 2024.

\end{thebibliography}
\bibliographystyle{iclr2026_conference}
\newpage
\appendix



\section{Hard Negative Statistics}
\label{sec:hard_negative_statistics}

We conduct a detailed analysis of the constructed hard negative datasets, covering the data sources, construction methods, the number of positive samples, the number of hard negatives per positive sample, and additional notes on the perturbation strategies. The statistics are summarized in Table~\ref{tab:hard-negatives}.

\begin{table*}[ht]
\centering
\small
\setlength{\tabcolsep}{3pt}
\resizebox{\linewidth}{!}{
\begin{tabular}{lcccl}
\toprule
\textbf{Data Type} & \textbf{Source \& Method} & \textbf{\#Positives} & \textbf{\#Hard Negatives} & \textbf{Notes} \\
\midrule
Retrieval‑based Negative & MAVIS + SimANS & 100K & 10-50 & Dense retrieval, high semantic similarity \\
Rule‑based Negative      & MAVIS + GLM‑4  & 100K & 10 & Shape / relation / value perturbations \\
Image‑based Negative     & New Collection + Gemini2.5 & 4K & 10 & Code‑driven geometric modifications \\
\bottomrule
\end{tabular}
}
\caption{Statistics of constructed hard‑negative samples. “\#Hard Negatives” indicates the number of negatives per positive sample.}
\label{tab:hard-negatives}
\end{table*}

\section{Examples of Image-based Negatives}
\label{sec:case of image_based negatives}
\begin{figure*}[!ht]
    \centering
    \small
    \includegraphics[width=0.95\linewidth]{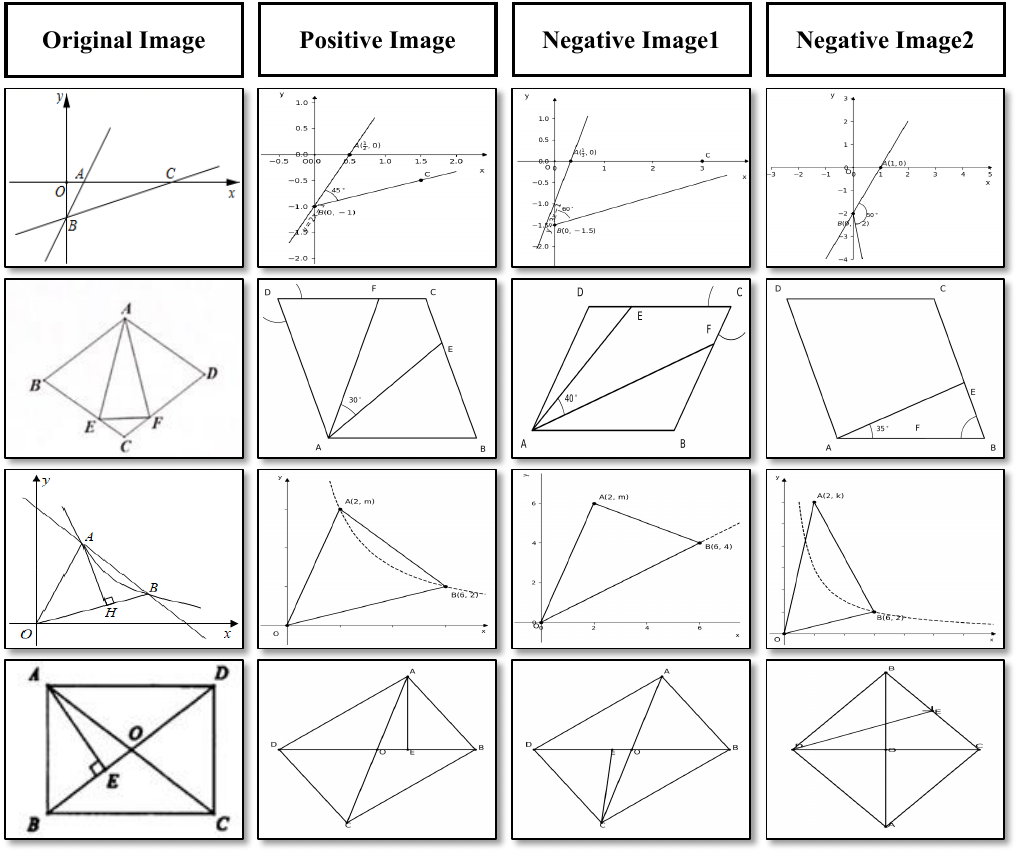}
    \caption{Comparison of generated geometric figures. The first column shows the original images, the second column presents the positive images, and the last two columns illustrate the constructed hard negatives.}
    \label{fig:image_compare}

\end{figure*}

Figure~\ref{fig:image_compare} presents some examples of image-based negatives, covering analytical geometry and planar geometry. Geometric elements include triangles, quadrilaterals, and various relational properties such as perpendicularity and intersection. Additionally, certain length/angle properties absent from the original diagrams are explicitly indicated. Specifically, the first column shows the original geometric images, and the second column depicts images generated from the given questions and the true answer, which, despite slight discrepancies, accurately capture the overall outlines of the original diagrams. The third and fourth columns display the constructed hard negative image based upon their corresponding negative captions and corresponding Python script—demonstrating notable similarities but clear and meaningful distinctions.

\section{Retrieval Experiments for Vision Encoder}
\label{sec:retieval}

In addition to ablation evaluation, retrieval performance offers a direct measure of image–text alignment, computed via similarity scores between paired inputs. As a result, the ranking position of the ground-truth positive sample reflects the quality of learned representations.
To assess retrieval capabilities, we evaluate the trained AltCLIP models using the \textit{Hit@1} metric, which measures whether the correct caption is ranked highest among a large pool of candidates. Specifically, we construct four evaluation sets of hard negatives, each containing 500 samples, corresponding to the following strategies:
\begin{itemize}
    \item \textbf{Random Negative}: Caption negatives randomly selected.  
    \item \textbf{Retrieval Negative}: Caption negatives  retrieved using the SimANS model, selecting the top 100 most similar captions to the positive sample.  
    \item \textbf{Rule-based Negative}: Caption negatives generated according to the rules defined in earlier section.  
    \item \textbf{Image-based Negative}: Image negatives generated from real-world geometric problems from new collection.
\end{itemize}
Table~\ref{tab:clip_variants} reports retrieval performance of various vision encoders trained with different types of negative samples in this experiment.

\begin{figure*}[!ht]
    \centering
    \small
    \includegraphics[width=0.95\linewidth, trim=0 10 0 0, clip]{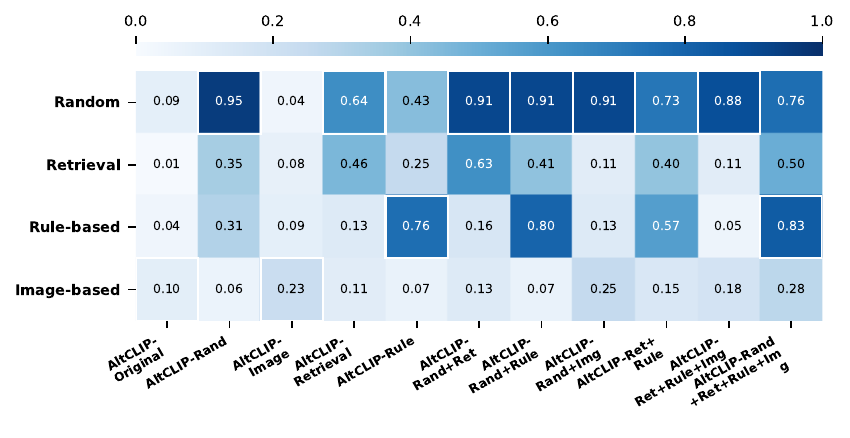}
\caption{Retrieval performance of AltCLIP variants trained with different strategies.}

    \label{fig:clip-performance}

\end{figure*}



\begin{table}[tb]
\centering
\small
\setlength{\tabcolsep}{4pt}
\begin{tabular}{lcccccl}
\toprule
\multirow{2}{*}{\textbf{AltCLIP Variant}} & 
\multicolumn{4}{c}{\#Negative Source (K)} & 
\multirow{2}{*}{\makecell{\textbf{Neg}\\\textbf{pos}}} & 
\multirow{2}{*}{\textbf{Train}} \\ 
\cmidrule(lr){2-5}
& \textsc{Rnd} & \textsc{Ret} & \textsc{Rule} & \textsc{Img} & & \\ 
\midrule
AltCLIP-Original    & --  & --  & --  & -- & -- & -- \\
AltCLIP-Rand        & 400 & --  & --  & -- & 10 & IB \\
AltCLIP-Retrieval   & --  & 100 & --  & -- & 10 & MM \\
AltCLIP-Rule        & --  & --  & 100 & -- & 10 & MM \\
AltCLIP-Image       & --  & --  & --  & 4  & 10 & MM \\
AltCLIP-Rand+Ret    & 400 & 100 & --  & -- & 10 & IB+MM \\
AltCLIP-Rand+Rule   & 400 & --  & 100 & -- & 10 & IB+MM \\
AltCLIP-Rand+Img    & 400 & --  & --  & 4  & 10 & IB+MM \\
AltCLIP-Ret+Rule    & --  & 100 & 100 & -- & 10 & MM \\
AltCLIP-Ret+Rule+Img& --  & 100 & 100 & 4  & 10 & MM \\
AltCLIP-All         & 400 & 100 & 100 & 4  & 10 & IB+MM \\
\bottomrule
\end{tabular}
\caption{Dataset composition for each \textbf{AltCLIP} vision encoder variant.  
Columns list the number of positive pairs (in thousands) drawn from four negative-source types:  
\textbf{\textsc{Rnd}} = random, \textbf{\textsc{Ret}} = retrieval, \textbf{\textsc{Rule}} = rule-based, \textbf{\textsc{Img}} = image-based.  
\emph{Neg/pos} indicates negatives-per-positive;  
training strategies: \textbf{IB} = In-batch, 
\textbf{MM} = MMCLIP, 
\textbf{IB+MM} = hybrid.
}
\label{tab:clip_variants}
\end{table}


The retrieval performance of different AltCLIP variants is summarized in Figure~\ref{fig:clip-performance}.
Under the image-based negative setting, AltCLIP-Image achieves the highest \textit{Hit@1} score (23\%), substantially outperforming all other variants, most of which remain below 15\%. This demonstrates the superior alignment learned from visually grounded negatives. However, the performance is still constrained, likely due to the limited size of the image-based dataset (4K samples), which may restrict the model’s capacity to generalize under this setting.
In retrieval- and rule-based evaluations, AltCLIP-Rand achieves the highest scores among single-source variants (64\% and 31\%, respectively), outperforming both AltCLIP-Retrieval and AltCLIP-Rule. This suggests that while domain-specific negatives help the model specialize, pre-training on random negatives provides broader generalization capabilities.
Combining random negatives with domain-specific ones further improves robustness. For instance, AltCLIP-Rand+Rule yields strong performance across retrieval (41\%) and rule-based (80\%) settings, while AltCLIP-Rand+Ret+Rule+Img consistently achieves top scores across all four settings. These results indicate that starting with random negatives and progressively incorporating hard negatives can effectively balance generalization and domain-specific precision in image–text alignment.

\section{Why Fewer Shot Image-Based Negatives Work: A Similarity and Distribution Study}
\label{sec:4kimage_based}

In above ablation study, we observe that only 4K image-based negatives generated  can match or even surpass the performance of 100K retrieval-based negative image from MAVIS. To better understand this phenomenon, we conduct two follow-up analyses.
\begin{table}[!tb]
\centering
\small
\setlength{\tabcolsep}{4pt}
\begin{tabular}{lcccc}
\toprule
\multirow{2}{*}{\textbf{Negative Type}} &
\multicolumn{4}{c}{\textbf{Max Cosine Similarity}} \\
\cmidrule(lr){2-5}
& $>0.90$ & $>0.85$ & $>0.70$ & $>0.60$ \\
\midrule
Image‑based \textit{vs.} SFT        & 20.89\% & 78.68\% & 98.43\% & 99.43\% \\
Retrieval‑based \textit{vs.} SFT    & 0.00\%  & 26.18\% & 98.00\% & 99.28\% \\
\bottomrule
\end{tabular}
\caption{Semantic similarity between hard negative captions and the 17K SFT dataset.  
Values denote the percentage of negative captions whose \emph{maximum} cosine similarity with any SFT question exceeds the specified threshold.}
\label{tab:neg_sim_sft}
\end{table}

First, we evaluate the semantic similarity between negative samples and the supervised fine-tuning  dataset. Specifically, we employ the sup-simcse-roberta-large model to encode (1) the captions of image-based negatives, (2) the top-10 retrieval-based negative captions, and (3) the questions from the 17K SFT dataset. We then compute the cosine similarity between each caption and the questions in SFT, and for each caption, we retain the maximum similarity score as its semantic alignment score.
Next, we calculate the proportion of captions whose maximum similarity exceeds various thresholds (0.90, 0.85, 0.70, and 0.60). As shown in Table~\ref{tab:neg_sim_sft}, image-based negatives exhibit substantially higher similarity to the SFT dataset, with 20.89\% exceeding 0.90 and 78.68\% exceeding 0.85. In contrast, retrieval-based negatives show much lower proportions, with only 0.00\% and 26.18\% exceeding these respective thresholds.
These results suggest that image-based negatives, which are derived from real-world problems, exhibit closer semantic alignment with downstream supervised data compared to artificially constructed  samples.

Next, we visualized the embeddings of both image-based negatives and retrieval-based negatives using t-SNE, reducing them to two dimensions, as shown in Figure~\ref{fig:tsne}. The visualization clearly shows a distinct separation between the two types of negative samples. We further evaluate the separation of the two negative types using K-means clustering on the embeddings of both image-based and retrieval-based negatives. We then calculated two separation metrics: K-means Separation Accuracy and Natural Separation Score. The results are as Table~\ref{tab:clustering_metrics}:
\begin{figure}[!ht]
    \centering
    \small
    \includegraphics[width=0.8\linewidth, trim=0 40 0 40, clip]{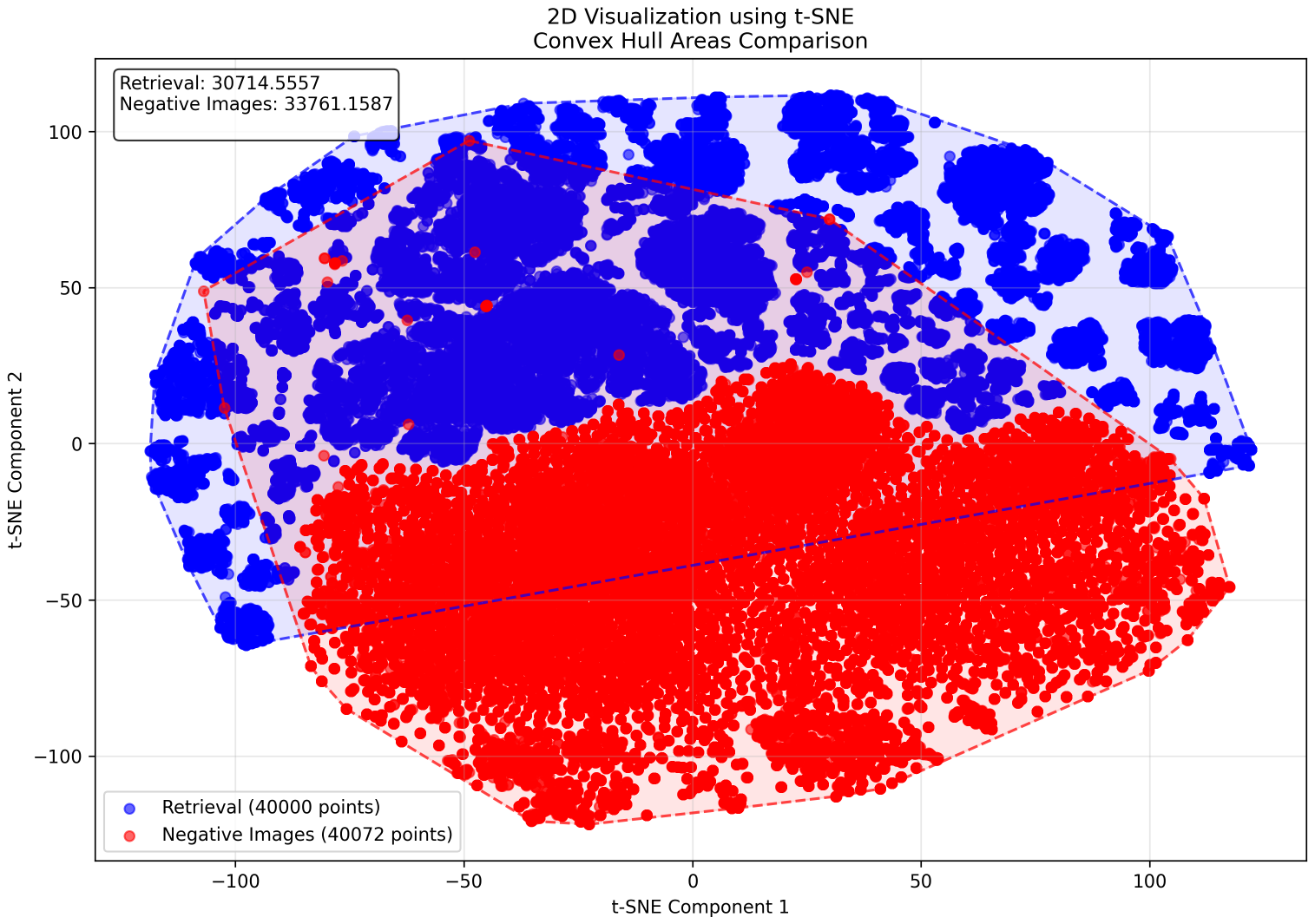}
    \caption{Comparison of GPT-4o and MMGeoLM in geometric problem-solving. Both models produce incorrect answers, but MMGeoLM's solution is closer to the True Answer.}
    \label{fig:tsne}
\end{figure}
\begin{table}[!htb]
\centering
\small
\setlength{\tabcolsep}{6pt}
\begin{tabular}{l c}
\toprule
\textbf{Metric} & \textbf{Value} \\
\midrule
K-means Separation Accuracy & 0.90 \\
Natural Separation Score     & 0.90 \\
\bottomrule
\end{tabular}
\caption{Clustering-based separation metrics between image-based and retrieval negative samples. Higher values indicate stronger distinguishability between the two groups in embedding space.}
\label{tab:clustering_metrics}
\end{table}

From Table~\ref{tab:clustering_metrics}, we observe that both the K-means Separation Accuracy and the Natural Separation Score reach 0.90, indicating a strong degree of separability between the two types of negatives. This suggests that the 4K image-based negatives are not only distinguishable from retrieval-based ones, but also semantically closer to downstream tasks. Consequently, their effectiveness stems from their higher alignment with real-world data, rather than relying on large-scale rule-based construction. In contrast, datasets built purely through human-making rules may require a substantially larger volume to achieve comparable performance, resulting in lower data efficiency.

\section{Training-Test Overlap and Data-Contamination Analysis}
\label{sec:training-test overlap}
Both the training and test datasets are drawn from genuine exam questions in an open‑ended format; thus, accidental duplicates could harm evaluation scores. To assess this risk, we embed every question with \textit{sup-simcse‑roberta‑large} and compute the cosine similarity between each test question and the entire training datasets.  For each test question we record its \emph{maximum} and \emph{top‑5} similarity scores.

\begin{figure*}[!ht]
    \centering
    \small
    \includegraphics[width=0.95\linewidth]{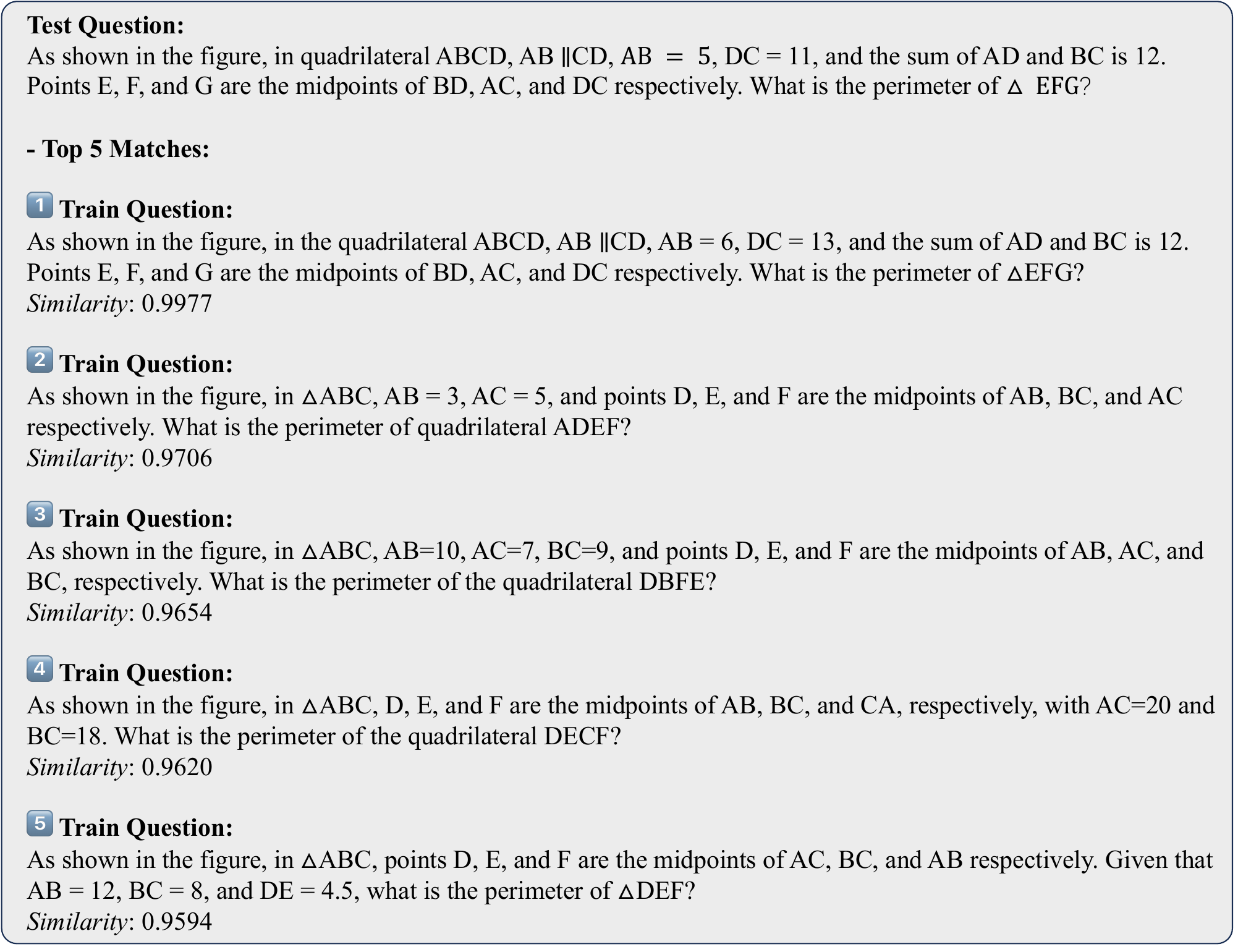}
\caption{Examples of top-5 most similar training–test question pairs used for contamination analysis. Each pair includes the test question, its top training matches, and the associated cosine similarity.}

    \label{fig:contam}

\end{figure*}

Figure~\ref{fig:contam} shows the examples of top 5  similar score pair .  For the top 1 score (\textit{sim}$=0.9977$), although their wording is close, a key numerical change (\,$AB=5, DC=11$ vs.\,$AB=6,DC=13$) yields a absolute different solution path. To ensure a clean separation between training and test sets, we remove from the training data any question whose similarity with any test question exceeds $0.995$. Table~\ref{tab:sim‑stats} summarizes the resulting overlap statistics. Under this filtering strategy, \textbf{100\%} of the test questions have no near-duplicate counterparts in the training data, and over \textbf{97\%} have a maximum similarity below $0.990$, demonstrating that our evaluation is not compromised by training leakage.
\begin{table}[!ht]
\centering
\footnotesize
\begin{tabular}{lcc}
\toprule
\textbf{Similarity Threshold} & \textbf{Percentage} \\
\midrule
$<0.995$  & 100\% \\
$<0.990$  & 97.28\% \\
$<0.980$ & 90.56\% \\
$<0.970$  & 81.26\% \\
\bottomrule
\end{tabular}
\caption{Proportion of test questions whose maximum similarity to the training set falls below various thresholds, after filtering.}
\label{tab:sim‑stats}
\end{table}

\section{Human Verification of Generated Captions}
\label{sec:hunman verification}
We randomly sampled 123 captions generated by the LLM call from the pipeline of image-based negatives. The sample size was chosen to cover diverse geometric configurations (triangles, quadrilaterals, circles, etc.). Each caption was independently evaluated by three annotators with a background in mathematics education at the college level. Annotators were asked to check whether every geometric element described in the caption (e.g., points, lines, angles, shapes) was consistent with the diagram rendered from code. The task was framed as a binary decision: correct (all elements accurately represented) vs. incorrect (at least one element misrepresented or missing).
All three annotators agreed on every instance, yielding 100\% accuracy with full inter-annotator agreement. Although the agreement was perfect on this sample, the evaluation was limited to 123 captions. Larger-scale or more diverse verification may reveal edge cases where captions misrepresent subtle geometric relations.

 \section{Prompts}
\label{sec:prompts}
\begin{figure}[!ht]
    \centering
    \small
    \includegraphics[width=0.95\linewidth]{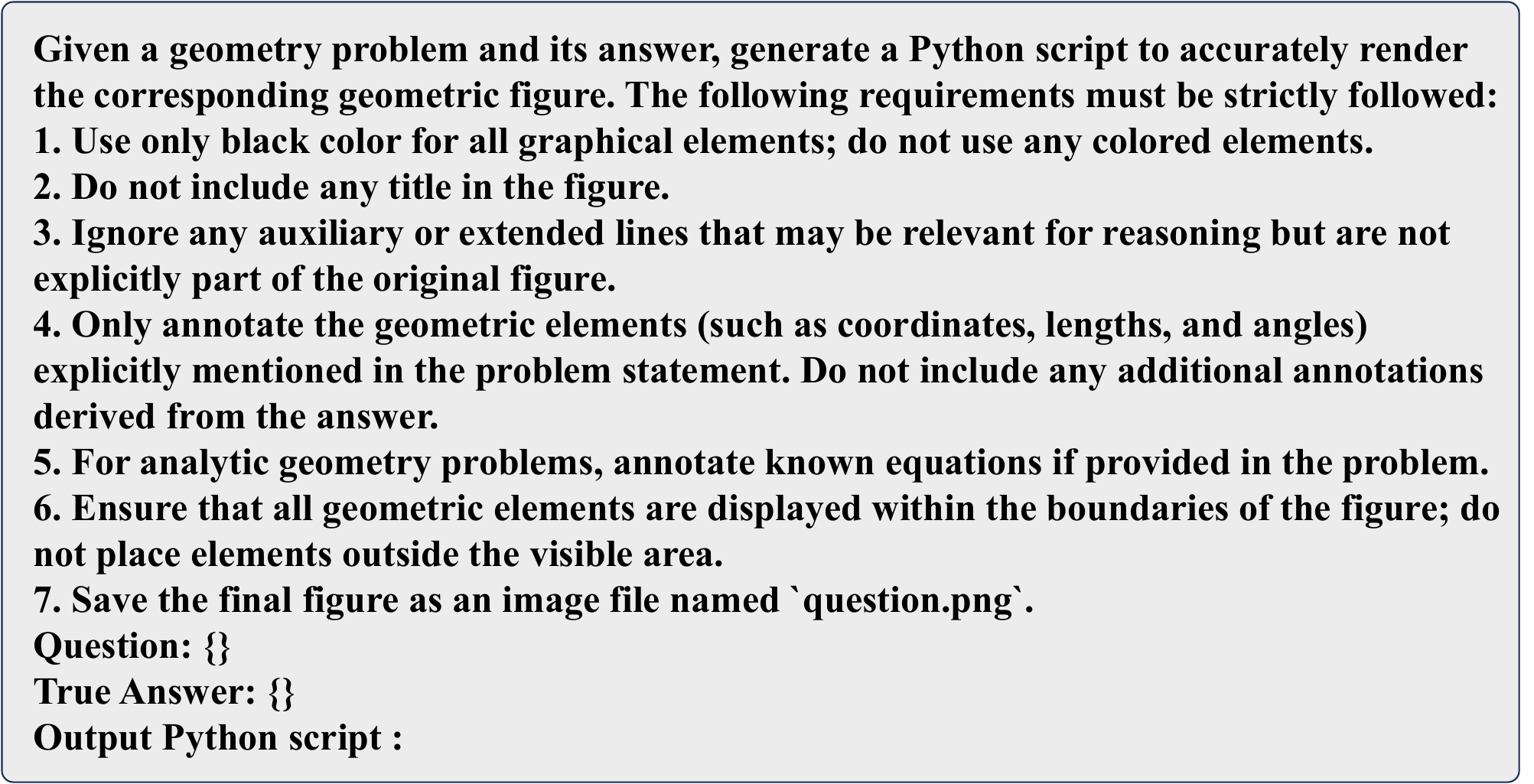}
\caption{Prompt for generating a geometric Python script from a given geometry problem and its solution. The resulting script is used to render the corresponding image.}
    \label{fig:python frompt}

\end{figure}

\begin{figure}[!ht]
    \centering
    \small
    \includegraphics[width=0.95\linewidth]{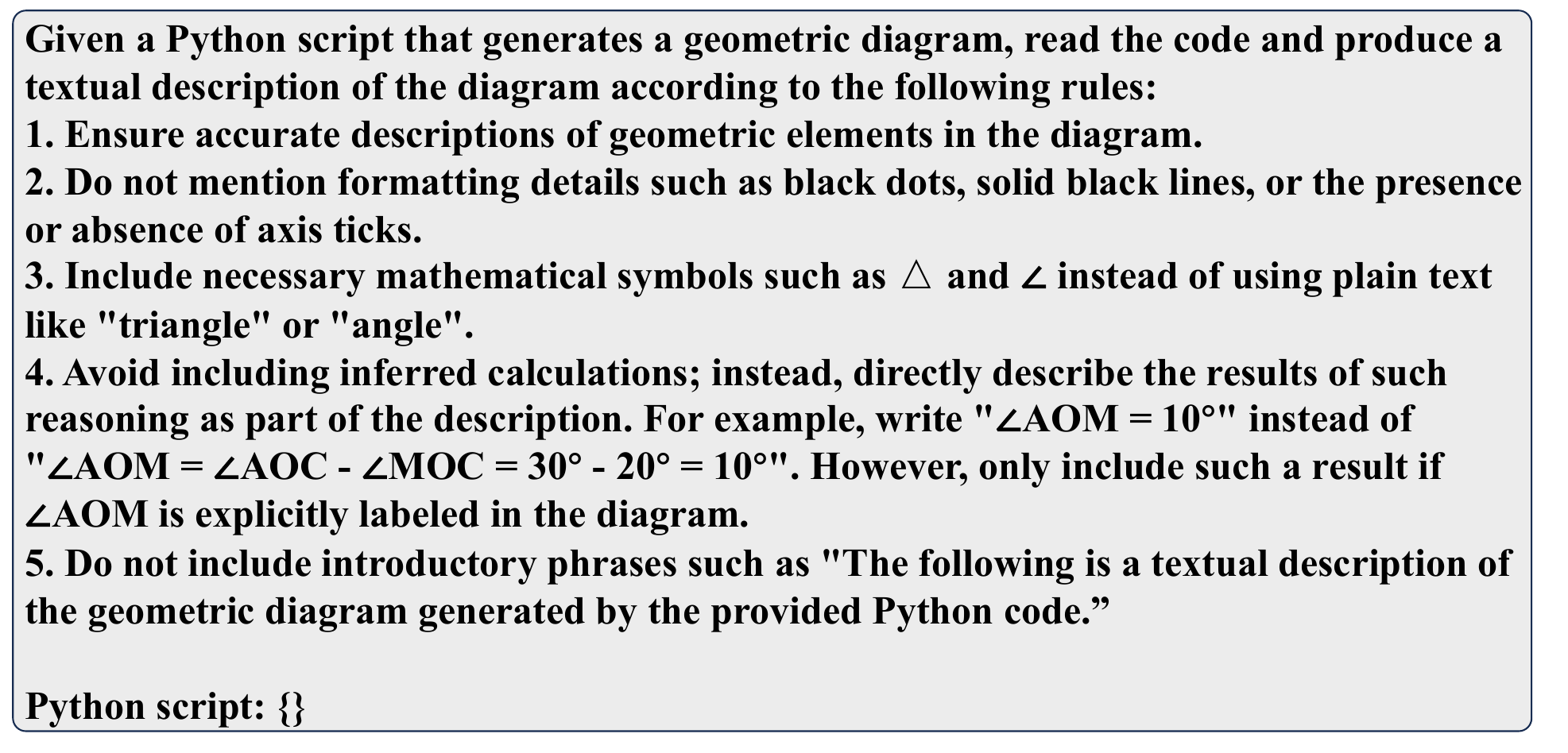}
\caption{Prompt for generating a positive geometric caption from a Python script. }
    \label{fig:positive caption}

\end{figure}

\begin{figure}[!ht]
    \centering
    \small
    \includegraphics[width=0.95\linewidth]{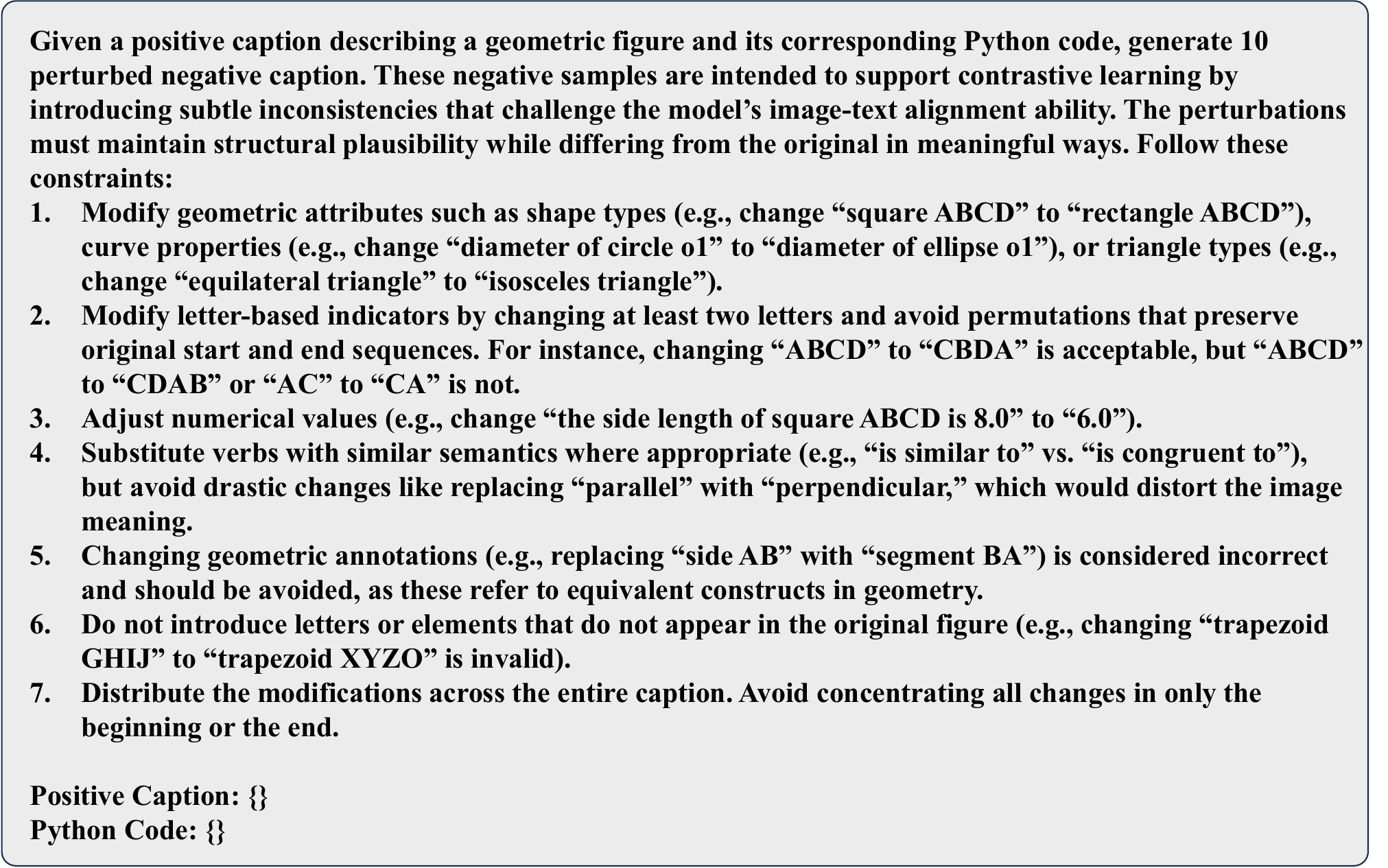}
\caption{Prompt for generating negative captions by introducing fine-grained semantic modifications to the positive caption.}

    \label{fig:negative caption}

\end{figure}

\begin{figure}[!ht]
    \centering
    \small
    \includegraphics[width=0.95\linewidth]{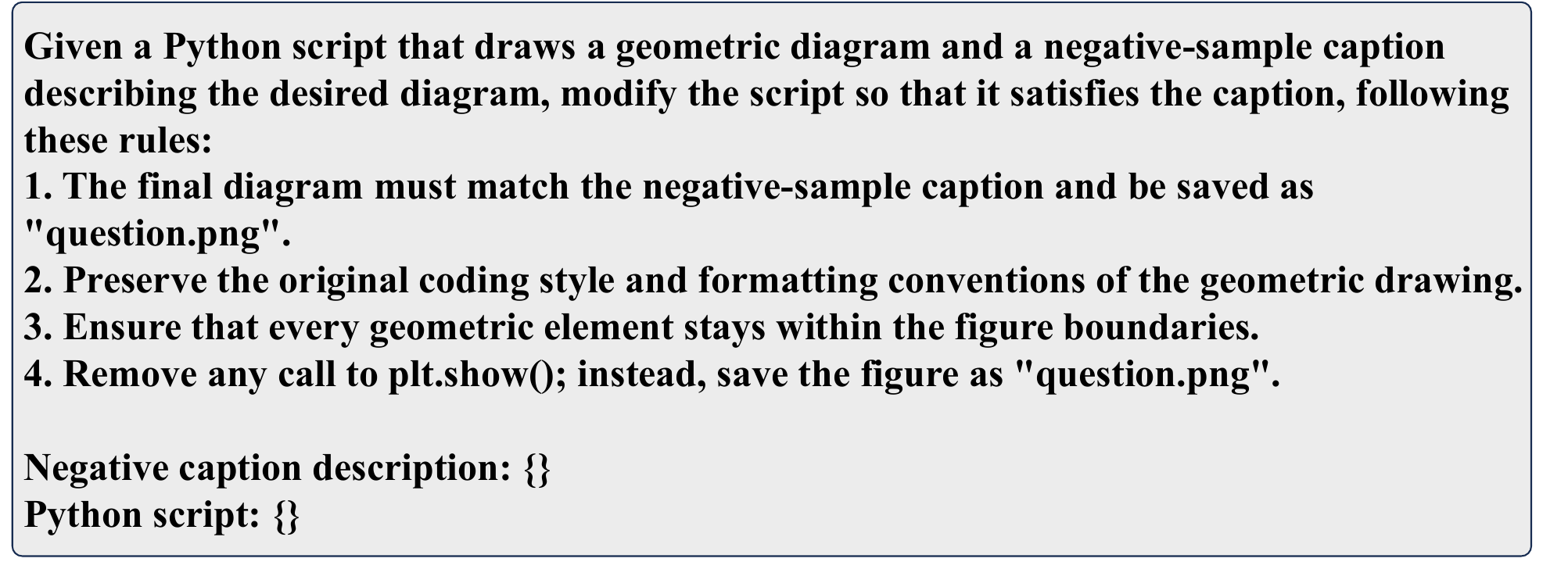}
\caption{Prompt for generating negative geometric images by modifying the original Python script based on the perturbed negative captions.}

    \label{fig:negative code}

\end{figure}

\begin{figure}[!ht]
    \centering
    \small
    \includegraphics[width=0.95\linewidth]{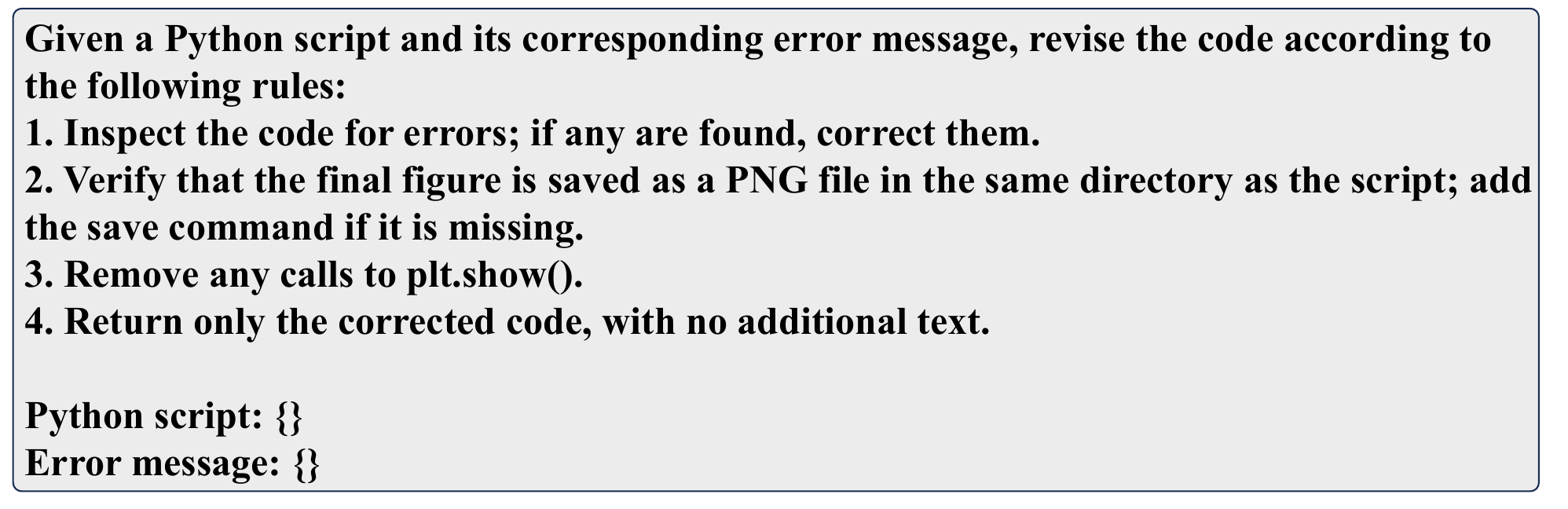}
\caption{Prompt for automatic correction of syntactic or semantic errors in Python scripts using an LLM. This ensures that all generated diagrams are valid and executable.}

    \label{fig:error fix}

\end{figure}

\begin{figure}[!ht]
    \centering
    \small
    \includegraphics[width=0.95\linewidth]{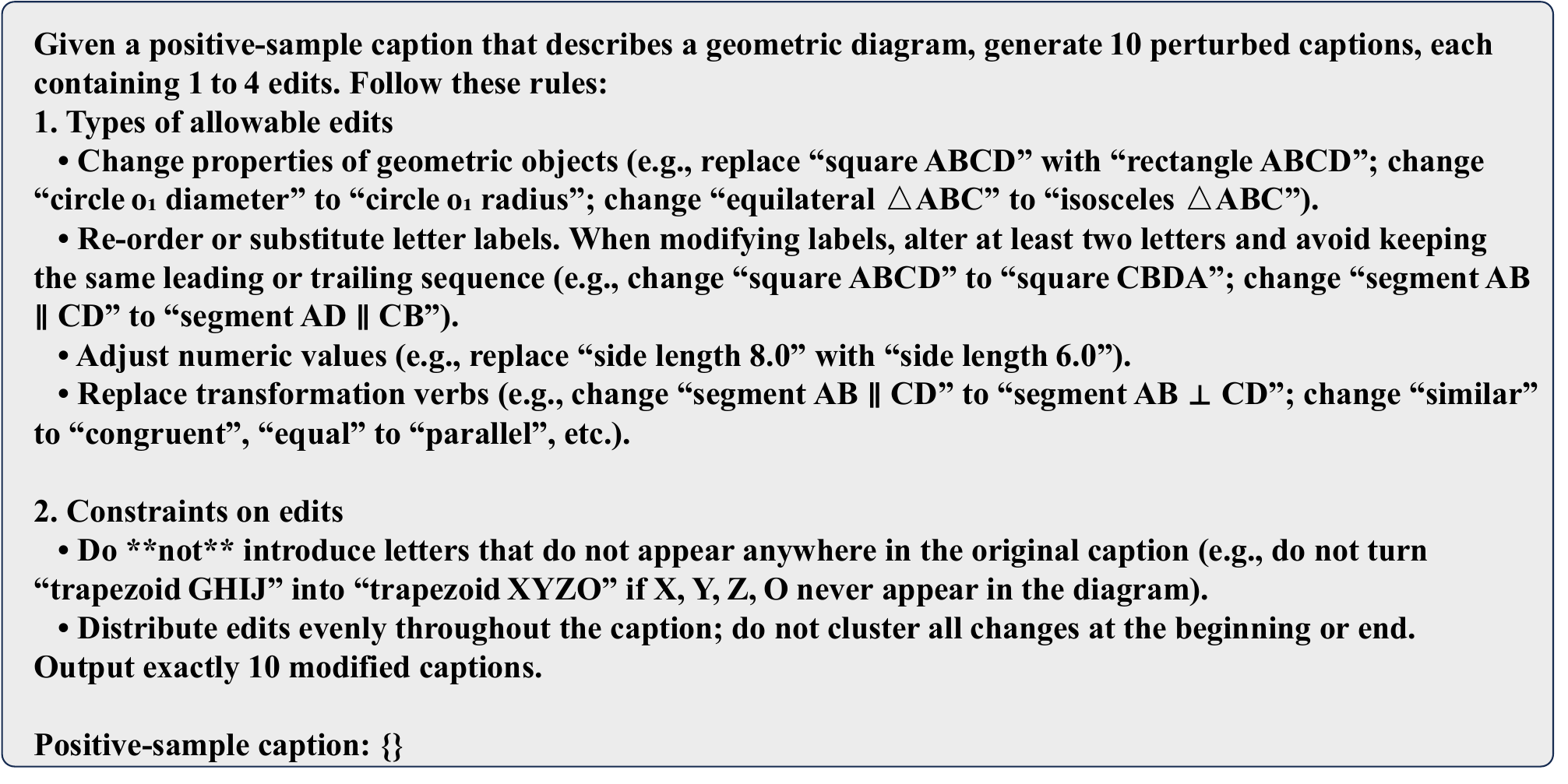}
\caption{Prompt for constructing rule-based text negatives through controlled perturbations of the original caption. }

    \label{fig:rule based}

\end{figure}

After multiple rounds of experimentation, we finalized six prompts for our data construction, as illustrated in Figures~\ref{fig:python frompt} to~\ref{fig:rule based}. Specifically, the prompt in Figure~\ref{fig:python frompt} input a textual geometry problem along with its answer and generates a corresponding Python script that renders the geometric image. Based on this script, we further generate a positive caption using the prompt shown in Figure~\ref{fig:positive caption}.
Directly perturbing the Python script alone to get image often leads to uncontrolled. Therefore, as shown in Figure~\ref{fig:negative caption}, we first generate negative captions by introducing subtle but semantically significant modifications to the positive caption. We then use these negative captions to  perturb  the original Python script, ultimately producing visually plausible but negative images, as shown in Figure~\ref{fig:negative code}.
Since the images are programmatically generated, errors in the scripts are sometimes inevitable. To ensure quality, we employ a LLM to automatically verify and correct any syntactic or semantic issues in the generated code by Figure~\ref{fig:error fix}.
For rule-based text negatives, we design specialized prompts to enforce targeted modifications. The corresponding prompt is illustrated in Figure~\ref{fig:rule based}, which enables controlled perturbations based on observed reasoning errors.


\section{Case Study of Problem Solving}
\label{sec:case sturdy}
\begin{figure*}[!ht]
    \centering
    \small
    \includegraphics[width=0.95\linewidth]{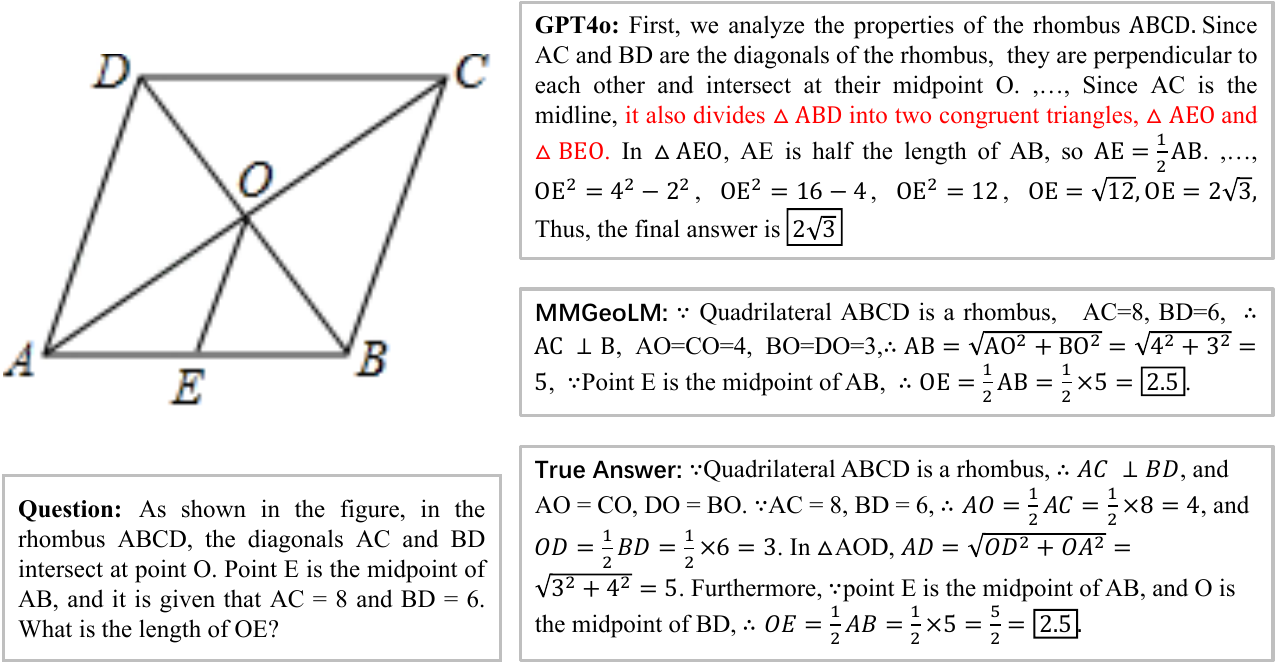}
    \caption{Comparison of GPT-4o and MMGeoLM in geometric problem-solving. MMGeoLM correctly answers the problem using a different solution compared to True Answer, while GPT-4o generates a non-existent element $\triangle ABD$.}
    \label{fig:case}
\end{figure*}
\begin{figure*}[!ht]
    \centering
    \small
    \includegraphics[width=0.8\linewidth]{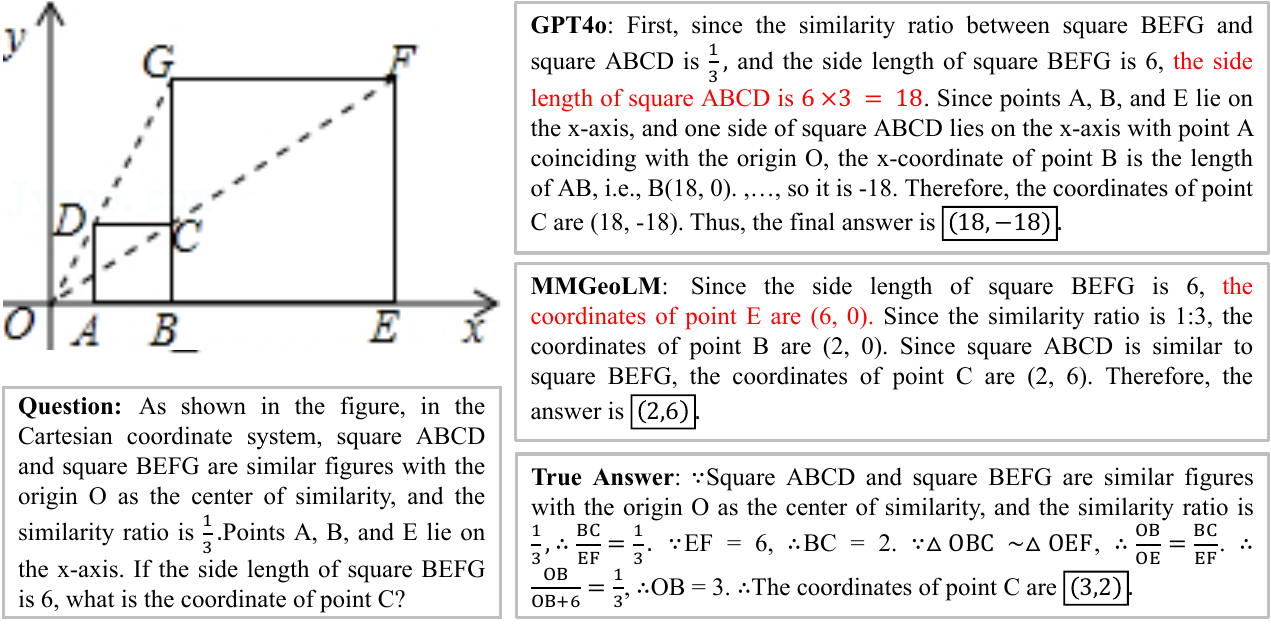}
    \caption{Comparison of GPT-4o and MMGeoLM in geometric problem-solving. Both models produce incorrect answers, but MMGeoLM's solution is closer to the True Answer.}
    \label{fig:case2}
\end{figure*}

In Figures~\ref{fig:case} and \ref{fig:case2}, we compare the solutions generated by GPT-4o and our MMGeoLM. In Figure~\ref{fig:case}, the red text highlights GPT-4o's misrecognition of two triangles in the image, leading to an incorrect final result. In contrast, MMGeoLM produces a correct solution, albeit through a reasoning process different from the ground-truth answer. This highlights MMGeoLM's capability to generate diverse problem-solving approaches.
Figure~\ref{fig:case2} illustrates errors in image-based reasoning for both MMGeoLM and GPT-4o. While both models make mistakes, MMGeoLM's reasoning aligns more closely with the correct solution, demonstrating improved mathematical reasoning. Overall, our trained MMGeoLM model demonstrates greater reasoning diversity and enhanced geometric problem-solving performance compared to current mainstream models.

\end{document}